\documentclass{IEEEtran}

\usepackage{graphicx}
\usepackage{amsmath}
\usepackage{multirow}
\usepackage{booktabs}
\usepackage{url}
\usepackage{subfigure}
\usepackage{xcolor}
\usepackage{slashbox}
\usepackage{array}
\usepackage{bm}
\ifCLASSINFOpdf
\else
\fi
\begin{document}
%
\title{Defect detection for patterned fabric images based on GHOG and low-rank decomposition}
%
%
%

\author{Chunlei Li(1),
        Guangshuai Gao(1),
        Zhoufeng Liu(1),
        Di Huang(2),
        Sheng Liu(1),
        Miao Yu(1)

        \IEEEauthorblockA{(1) School of Electronic and Information Engineering, Zhongyuan University of Technology,
ZhengZhou 450007, China\\}
 \IEEEauthorblockA{(2) School of Computer Science and Engineering, Beihang University, BeiJing, 100191, China}
}

%
%

\markboth{}%
{Shell \MakeLowercase{\textit{et al.}}: Bare Demo of IEEEtran.cls for IEEE Journals}
%



\maketitle

\begin{abstract}
In order to accurately detect defects in patterned fabric images, a novel detection algorithm based on Gabor-HOG (GHOG) and low-rank decomposition is proposed in this paper. Defect-free pattern fabric images have the specified direction, while defects damage their regularity of direction. Therefore, a direction-aware descriptor is designed, denoted as GHOG, a combination of Gabor and HOG, which is extremely valuable for localizing the defect region. Upon devising a powerful directional descriptor, an efficient low-rank decomposition model is constructed to divide the matrix generated by the directional feature extracted from image blocks into a low-rank matrix (background information) and a sparse matrix (defect information). A nonconvex log det($\cdot$) as a smooth surrogate function for the rank instead of the nuclear norm is also exploited to improve the efficiency of the low-rank model. Moreover, the computational efficiency is further improved by utilizing the alternative direction method of multipliers (ADMM). Thereafter, the saliency map generated by the sparse matrix is segmented via the optimal threshold algorithm to locate the defect regions. Experimental results show that the proposed method can effectively detect patterned fabric defects and outperform the state-of-the-art methods.
\end{abstract}


%
\IEEEpeerreviewmaketitle

\section{Introduction}
Fabric defect detection always plays a key role in the quality control of textile industry. Currently, it is mainly performed visually by skilled workers. However, its reliability is restricted by eye fatigue and human errors. An automated detection system based on machine vision can provide a promising solution that not only minimizes labor costs, but will also improve accuracy and efficiency. Moreover, an automated system is better equipped to deal with different kinds of fabric patterns, from the non-motif pattern (plain and twill fabrics, as shown in Fig.1 (a)) to the motif pattern (star-, box-, and dot-patterned fabrics, as shown in Fig.1 (b-d)).

Most existing defect detection methods focus on simple plain and twill fabrics, which can be classified into four categories: statistical method [1], frequency analysis method [2], model method [3], and dictionary learning method [4].

The aforementioned defect detection methods achieve high detection accuracy. However, because of the complexity and sophisticated design on patterned fabric, these proposed methods cannot be extended to detect patterned fabric defects. Moreover, few studies have been conducted on patterned fabric so far. In this paper, our research is focused on defect detection in patterned fabric.

\begin{figure}[ht]
\centering
\subfigure[]{\includegraphics[width= 0.1\textwidth]{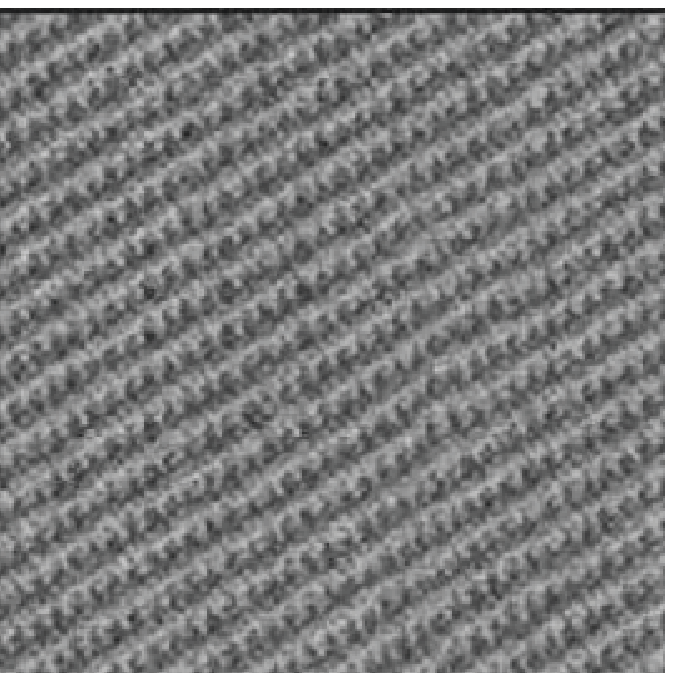}}\
\subfigure[]{\includegraphics[width= 0.1\textwidth]{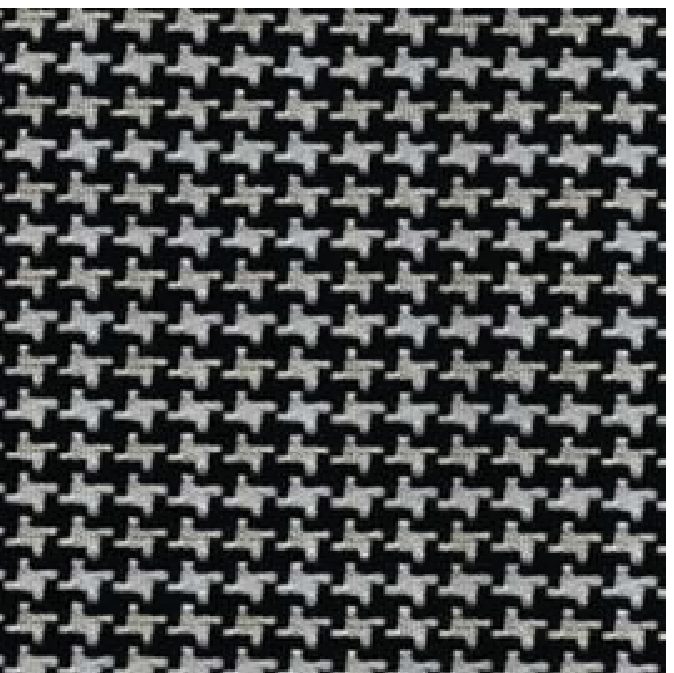}}\
\subfigure[]{\includegraphics[width= 0.1\textwidth]{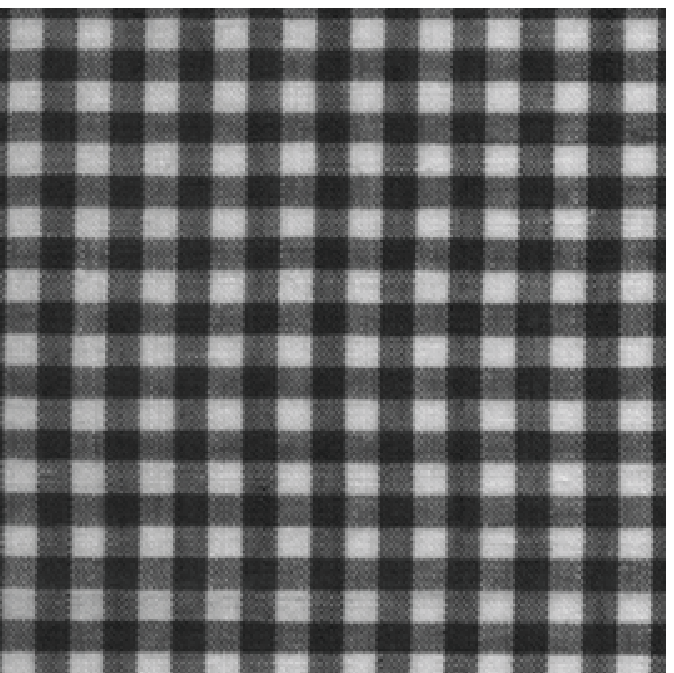}}\
\subfigure[]{\includegraphics[width= 0.1\textwidth]{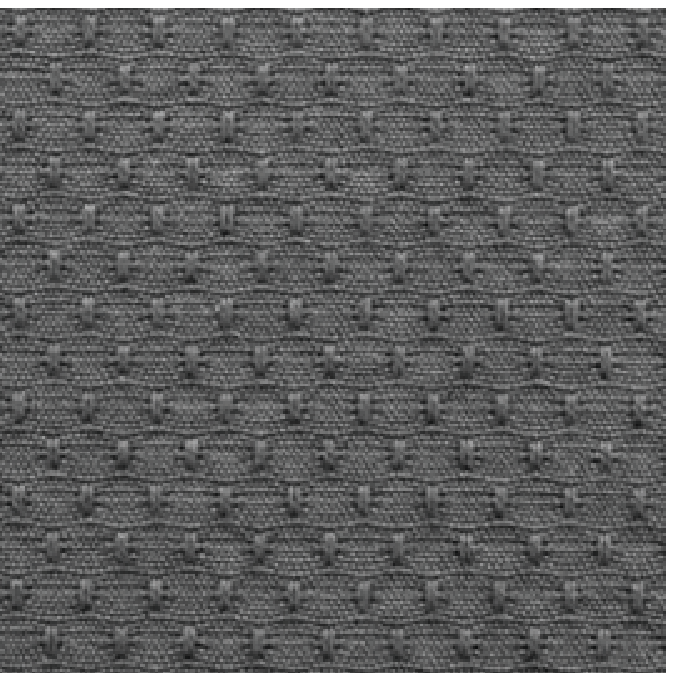}}\
  \caption{(a) Plain and twill. (b) Star-patterned fabric.(c) Box-patterned fabric. (d) Dot-patterned fabric.}
\end{figure}

The patterned fabrics are defined as fabrics with repetitive patterned units in their designs. Even within the same class of 'patterned' fabrics, there are still many categories, and the pattern sizes are different. Therefore, patterned fabric defect detection is a challenging task. Traditional methods cannot efficiently recognize the normal pattern; furthermore, they fail to localize the defect region. Some methods devised for pattern fabric defect detection, such as the ELO rating (ER) method [31], and wavelet-preprocessing golden image subtraction (WGIS) [21], et al.,which are performed in a supervised approach, require non-defective samples; moreover, the detection accuracy of these methods depends on precise alignment and choosing a suitable template.

Visual attention mechanism can enable machine and biological vision systems to quickly find the most salient regions or objects from a scene [5]. Generally, pattern fabric images always exhibit a high periodic texture among sub-patterns. However, the defect will disrupt their periodicity (regularity), resulting in the defects outstanding from homogeneous background. Therefore, visual saliency models provide a promising method for pattern fabric defect detection.

Many saliency models have been proposed to detect salient objects [5]. A representative series of papers are based on the low-rank matrix decomposition (LR) theory [6,7]. The background usually lies in a low-dimensional subspace, while the objects that are different from the background can be considered as noises or errors. Therefore, these methods divide an image into a low-rank matrix plus a sparse matrix in a learned feature space, where the low-rank matrix represents the background regions, and the sparse matrix indicates the salient object regions.

For different kinds of patterned fabrics, the background of the fabric has a visually homogeneous texture, and any defects stand out from the background. Compared with object detection in a natural scene, patterned fabric defect detection better fits the low-rank decomposition model.

However, using low-rank decomposition directly in a color space or in some other feature space of images does not deal with the task of pattern fabric defect detection with any sort of efficiency. This is because sub-parts have different colors or other features locally, but globally they belong to the normal texture. Therefore, a new powerful descriptor should be proposed to efficiently characterize the fabric texture, and should be required to possess the following attributes: (1) defect regions should have a totally different feature descriptor compared with the background; (2) the background should have a similar feature descriptor (that can be easily regarded as the low-rank part).

Defect-free pattern fabric images have a specified direction, while defects damage their regularity of direction. Therefore, a direction-aware descriptor can better represent the fabric feature, which is extremely valuable for separating a salient defect from a non-salient background.

Therefore, in this paper, a direction-aware descriptor, denoted as GHOG, is designed to be a combination of Gabor and HOG. Upon devising a powerful directional descriptor, an efficient low-rank decomposition model is constructed to divide the matrix generated by the directional feature extracted from image blocks into a low-rank matrix (background information) and a sparse matrix (defect information). Moreover, we also propose the use of a non-convex log det(¡¤) as a smooth surrogate function for the rank as opposed to the nuclear norm in order to improve the model's efficiency. Finally, the saliency map generated by the sparse matrix is segmented to locate the defect regions.

The reminder of this paper is organized as follows. Section 2 introduces the related works of fabric defect detection. In Section 3, we focus on the proposed algorithm and its specific procedures. Section 4 evaluates the performance of the proposed algorithm and compares it with other state-of-the-art methods. Finally, we conclude the paper in Section 5.

\section{Related works}
Defect detection plays an important role in the fabric quality control process. Many different fabric defect detection methods have been proposed to solve this problem, and are generally used to detect defects in plain and twill fabrics. These methods are divided into the following categories: spatial statistical analysis, spectral analysis, model-based methods, and dictionary learning. Spatial statistical methods detect defects by calculating gray values contrasted with their surroundings, including histogram character analysis [8], morphology [9], local contrast enhancement [10], and the fractal method [11]. The detection results of these methods depend largely on the size of a selected window and its discrimination threshold; it is difficult to detect smaller sizes defects for them. Additionally, these methods cannot effectively exploit the image's global information, and are always influenced by noise.

Spectral analysis methods transform the image to the spectral domain by choosing a suitable orthogonal basis, which can make better use of the image's global information detect defects. These methods include the Fourier transform (FT) [12], the Gabor transform [13] and the orthogonal wavelet transform [14]. However, these methods always have high computational complexity and poor detection performance for the complex texture fabric images.

Model-based methods first extract image texture features through modeling and parameter estimation techniques. Defect detection is realized by discriminating whether the test image conforms with the normal texture model. Existing methods include the Gaussian-Markov random field (GMRF) [15], and the Gaussian mixture model (GMM) [16, 17]. These methods have obtained satisfactory detection performance; however, they usually share a high computational complexity, and these methods cannot efficiently detect smaller size defects.

Dictionary learning-based methods learn a dictionary of training images or test images, and then reconstruct the defect-free fabric image; thereafter, defect detection can be realized by subtracting the recovered image from the test image [18, 19]. In a different way, dictionary learning based methods also reduce the dimension of an image block by projecting the image block into a dictionary learning from reference image, then the SVDD is adopted to discriminate whether an image block is a defect block [20]. However, these methods are unable to achieve ideal detection performance because the reconstructed image by the dictionary learning from themselves may exist some defects, or the self-adaptability of these methods are reduced if the dictionary learns from the reference images.

Regarding complicated patterned fabrics, several methods have been recently published, such as the wavelet-preprocessing golden image subtraction method (WGIS) [21], the Bollinger band method (BB) [22], the regular band method (RB) [23], template matching for discrepancy measures (TMPM) [24], the pattern matching and subtraction approach [25-28], the Hash function method [29], and the regularity and local orientation (RLO) method [30].

The WGIS method utilized a golden image to perform a moving subtraction of each pixel along each row of every wavelet-pre-processed tested image. The BB and RB methods, designed by different combination of moving averages and standard deviations, utilized the regularity property of a patterned texture to carry out on dot-, box-and star-patterned fabrics. The TMPM method used a golden image-like approach to exploit a discrepancy measure as a fitness function to detect defectson patterned textures. The patterned matching and subtraction method performs a point-to-point comparison, which is inherently sensitive to image noise, misalignment and distortion. The hash function method utilizes the offset properties between defect-free and regular patterns to detect the defects; it is fast but is also sensitive to noise and unable to show the shape of any defects after segmentation. The ELO rating (ER) method [31], is a method in which the detection of fabric defects is similar to carrying out fair matches in the spirit of good sportsmanship. However, this method depends on partition size, the number of randomly located partitions, $w$-variable and constant $K$.

The above pattern fabric defect detection methods adopt traditional approaches to characterize the fabric texture, such as wavelet transform, Gabor transform, average value, standard deviation and regular bands. They devised the feature descriptors, but did not consider the characteristics of the pattern fabric image. On the other hand, most of the complicated pattern fabric defect detection methods adopted template-matching technology to localize the defect; they are performed in a supervised approach. The detection accuracy depends on precise alignment and choosing a suitable template. Therefore, in this paper, a powerful direction-aware descriptor was designed, denoted as GHOG. The descriptor considers the characteristic of the patterned fabric image, and a low-rank decomposition model that can better fit the defect detection problem is adopted to separate the salient defect from the non-salient background.

\section{The proposed algorithm}
Defect-free pattern fabric images have a specified direction, while defects damage their regularity of direction. Moreover, compared with the object detection in a natural scene, patterned fabric defect detection can better fit a low-rank decomposition model. Therefore, our paper proposes a novel method of patterned fabric defect detection, based on GHOG and low-rank decomposition, and it includes the following four steps: 1) GHOG feature extraction; 2) low-rank decomposition model construction; 3) optimal solution of the model; 4) the generation and segmentation of the saliency map.

\begin{figure*}
  \centering
  \includegraphics[width=0.9\textwidth]{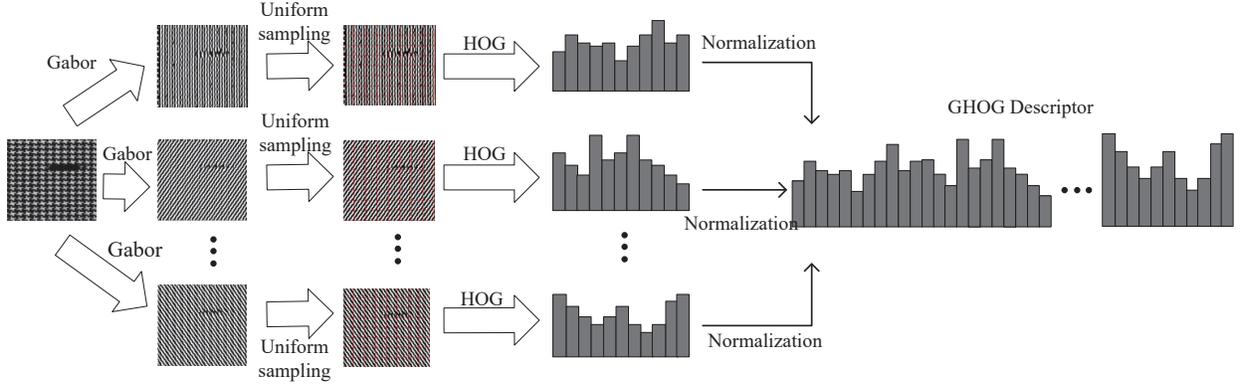}\\
  \caption{Construction process of the proposed GHOG descriptor}
\end{figure*}

\subsection{ GHOG feature extraction}
Highly effective fabric defect detection algorithms should resort to meaningful and powerful feature descriptors to facilitate uniqueness measurements and warrant sufficiently discriminative capabilities. Defect-free patterned fabric images have a specified direction, while defects damage their regularity of direction. Therefore, a direction-aware descriptor can better represent the fabric feature.

Due to the outstanding mathematical properties of the Gabor filter and its analogy to human visual mechanisms, the two-dimensional (2D) Gabor filter has been widely applied in the texture analysis. In this paper, a bank of Gabor directional filters have been adopted to extract directional information, and generate the directional Gabor filtered maps.

In addition, the gradient space of an image provides a measure of change in intensity over the pixels, as opposed to the absolute values of the pixels' intensity values. Although the existence of defects destroys the regularity of a patterned texture, it will have a more enhanced effect in gradient space domain than in image space domain. As a consequence, features extracted from gradient space domain of a defective periodic block will differ significantly to those extracted from a defective-free block. Considering that the Histogram of Oriented Gradients (HOG) features are extracted from generated Gabor filtered maps in quantized directions, a novel Gabor-HOG (GHOG) feature description is proposed to greatly reduce the feature dimension and to capture the directional feature. The construction process of the proposed GHOG descriptor is shown in Figure.2, and the specific procedures are described as follows:

1) \textbf{Gabor directional filtered map generation.} The complex function expression of a 2-D Gabor filter is described as
follows [34]:
\begin{equation}
g\left( {x,y;\lambda ,\psi ,\sigma ,\gamma } \right) = \exp \left( { - \frac{{x'^2  + \gamma ^2 y'^2 }}{{2\sigma ^2 }}} \right)\exp \left( {i\left( {2\pi \frac{{x'}}{\lambda } + \psi } \right)} \right)
\end{equation}
Its real part can be formulated as:
\begin{equation}
g\left( {x,y;\lambda ,\psi ,\sigma ,\gamma } \right) = \exp \left( { - \frac{{x'^2  + \gamma ^2 y'^2 }}{{2\sigma ^2 }}} \right)\cos \left( {2\pi \frac{{x'}}{\lambda } + \psi } \right)
\end{equation}
And its imaginary part is described as:
\begin{equation}
g\left( {x,y;\lambda ,\psi ,\sigma ,\gamma } \right) = \exp \left( { - \frac{{x'^2  + \gamma ^2 y'^2 }}{{2\sigma ^2 }}} \right)\sin \left( {2\pi \frac{{x'}}{\lambda } + \psi } \right)
\end{equation}
where $x' = x\cos \theta  + y\sin \theta$, $y' =  - x\sin \theta  + y\cos \theta$, $\lambda$ is wave length, which is always greater than or equal to 2, but not more than 1/5 of the size of the input image; $\theta$ represents directions, their values ranging from 0 to $2\pi$; $\psi$ indicates phase shift, whose range is from $-\pi$ to $\pi$, 0 and $\pi$ correspond to center-on and center-off functions, respectively, while $- \pi /2$ and $\pi /2$ correspond to anti-symmetric function; $\gamma$ is the length-width ratio, also known as the ratio of the vertical and the horizontal, which determines the ellipticity of the Gabor function; if $\gamma  = 1$, the shape is round, if $\gamma  < 1$ , shape stretches along the direction of parallel stripes, in this paper, $\gamma$ is set to 0.5; $\sigma$ is the Gaussian factor standard deviation of the Gabor function.The value of $\sigma$ cannot be preset directly, its change just depends on the variation of the bandwidth $b$. $b$ must be a positive constant, which is related to the ratio of $\sigma /\lambda$, we usually set it as 1. Then the relationship of $\sigma$ and $\lambda$ is $\sigma  = 0.56\lambda$. In this paper, empirically, we choose eight orientations with one scale to filter the patterned fabric image, and accordingly generate eight \textbf{directional filtered maps} $
I_{G_o }$ ($o$=1,2,...,\emph{N}, \emph{N}=8 here) that capture the directional features.

2) \textbf{Uniformly sampling for the Gabor filtered maps.} Once these directional filtered maps of all quantized directions are obtained, they are exploited as the inputs for the next computing histogram of orientated gradients features over the same image region. For each generated Gabor filtered map , the size of which is the same as the given original image, is equally decomposed into segments $\left\{ {I_{G_o }^i } \right\}_{i = 1,2...K}$ with sizes of $N_{\rm{b}}  \times N_{\rm{b}}$ ; where $o$ indicts the orientations, and $o$=1,2,...8, $K$ is the number of segments, $N_b$ equals 16 in this paper.

3) \textbf{ HOG feature extraction.}  For each segment $I_{G_o } ^i$, the HOG is similar to the method described in [35]. The detail procedures are as follows: 	

First of all, normalize the input image block $I_{G_{\rm{i}} } ^j$ obtained by gamma rectification, and it is described as follows.
\begin{equation}
\emph{H}_o (x,y) = (\emph{I}_{G_o }^i )^{gamma}
\end{equation}
In this paper, gamma is set to 1/2.

Calculate the gradient of image pixels along the horizontal and vertical directions:
\begin{equation}
mag_o (x,y) = \sqrt {\left( {\frac{{\partial H_o (x,y)}}{{\partial x}}} \right)^2  + \left( {\frac{{\partial H_o (x,y)}}{{\partial y}}} \right)^2 }
\end{equation}

\begin{equation}
\theta _o (x,y) = \arctan (\frac{{\partial H_o (x,y)}}{{\partial y}}/\frac{{\partial H_o (x,y)}}{{\partial x}})
\end{equation}

\begin{equation}
\frac{{\partial H_o \left( {x,y} \right)}}{{\partial x}} = H_o \left( {x + 1,y} \right) - H_o \left( {x - 1,y} \right)
\end{equation}

\begin{equation}
\frac{{\partial H_o \left( {x,y} \right)}}{{\partial y}} = H_o \left( {x,y + 1} \right) - H_o \left( {x,y - 1} \right)
\end{equation}

Then each orientation is mapped to the range of $[0,2\pi ]$ from that of $[ - \pi /2,\pi /2]$, which keeps consistent with the number of the Gabor filter maps. After quantization, the entry $n_o$ of each orientation $\theta _o$ is computed as follows:
\begin{equation}
n_o (x,y) = \bmod (\left\lfloor {\frac{{\theta _o (x,y)}}{{2\pi /N}} + \frac{1}{2}} \right\rfloor ,N),{\kern 1pt} {\kern 1pt} {\kern 1pt} {\kern 1pt} o = 1,2,...,N
\end{equation}

The histogram of orientation gradients features, $h_{oi}$, is constructed as (10) by accumulating the gradient magnitude $mag_o$ of all the pixels with the same quantized orientation entry $n_o$.
\begin{equation}
h_{oi} (j) = \sum {f(n_o (x,y) = j)} *mag_o (x,y)
\end{equation}
where \emph{j} =0, 1,..., \emph{N}-1; \emph{o}=1,2,...,\emph{N}; \emph{i} =1,2,...,\emph{K}.
\begin{equation}
f(x) = \left\{ \begin{array}{l}
 1,{\kern 1pt} {\kern 1pt} {\kern 1pt} {\kern 1pt} {\kern 1pt} {\kern 1pt} {\kern 1pt} {\kern 1pt} {\kern 1pt} {\kern 1pt} {\kern 1pt} {\kern 1pt} {\kern 1pt} {\kern 1pt} {\kern 1pt} if{\kern 1pt} {\kern 1pt} {\kern 1pt} x{\kern 1pt} {\kern 1pt} {\kern 1pt} is{\kern 1pt} {\kern 1pt} {\kern 1pt} true \\
 0,{\kern 1pt} {\kern 1pt} {\kern 1pt} {\kern 1pt} {\kern 1pt} {\kern 1pt} {\kern 1pt} {\kern 1pt} {\kern 1pt} {\kern 1pt} {\kern 1pt} {\kern 1pt} {\kern 1pt} {\kern 1pt} otherwise \\
 \end{array} \right.
\end{equation}

Then, for each Gabor filtered maps $I_{G_o }$, its HOG features $h_o$ is generated by concatenating all the histograms of all the segments:
\begin{equation}
h_o  = [h_{o1} ,h_{o2} ,h_{o3} ,...,h_{oK} ]^T
\end{equation}

The final GHOG descriptor is generated by concatenating all \emph{N} histogram of HOG features of all segments as (13). Each histogram $h_o$ is normalized to a unit norm vector $
\mathop {h_{o_K } }\limits^ \wedge$ before the concatenation.
\begin{equation}
GHOG_K  = \left[ {\mathop {h_{1K} }\limits^ \wedge  ,\mathop {h_{2K} }\limits^ \wedge  ,...,\mathop {h_{NK} }\limits^ \wedge  } \right]^T
\end{equation}

We define a feature matrix $\textbf{F}$ as the final GHOG descriptors to represent the information of the entire image.

\begin{equation}
\textbf{F} = [GHOG_1 ,GHOG_2 ,...,GHOG_K ]
\end{equation}

\subsection{Low-rank decomposition model}
Low-rank decomposition and subspace recovery have been widely used for detection and recognition. This is because the background of an image always lies in a low-dimensional subspace, while the objects to be detected are different from the background can be considered as either noise or errors. Therefore, these methods divide an image into a low-rank matrix, plus a sparse matrix in a learned feature space, where the low-rank matrix represents the background regions, and the sparse matrix indicates the salient object regions. For different kinds of patterns, the fabric's background has a visually homogeneous texture, and the defects stand out. Compared with object detection in a natural scene, the patterned fabric defect detection better fits the low-rank decomposition model. Therefore, in this paper, the low-rank decomposition model is proposed to detect the defect region.

For the aforementioned generated feature matrix $F$, a low-rank decomposition model is constructed as follows:

\begin{equation}
\begin{split}
 \left( {L^* ,S^* } \right) = \arg {\kern 1pt} {\kern 1pt} {\kern 1pt} \mathop {\min }\limits_{\left( {L,S} \right)} {\kern 1pt} {\kern 1pt} {\kern 1pt} {\kern 1pt} \left( {rank{\kern 1pt} {\kern 1pt} {\kern 1pt} \left( L \right) + \lambda \left\| S \right\|_0 {\kern 1pt} } \right){\kern 1pt} {\kern 1pt}  \\
 {\kern 1pt} {\kern 1pt} {\kern 1pt} {\kern 1pt} {\kern 1pt} {\kern 1pt} {\kern 1pt} {\kern 1pt} {\kern 1pt} {\kern 1pt} {\kern 1pt} {\kern 1pt} {\kern 1pt} {\kern 1pt} {\kern 1pt} {\kern 1pt} {\kern 1pt} {\kern 1pt} {\kern 1pt} {\kern 1pt} {\kern 1pt} {\kern 1pt} {\kern 1pt} {\kern 1pt} {\kern 1pt} {\kern 1pt} {\kern 1pt} {\kern 1pt} {\kern 1pt} {\kern 1pt} {\kern 1pt} {\kern 1pt} {\kern 1pt} {\kern 1pt} {\kern 1pt} {\kern 1pt} {\kern 1pt} {\kern 1pt} {\kern 1pt} {\kern 1pt} {\kern 1pt} {\kern 1pt} {\kern 1pt} {\kern 1pt} {\kern 1pt} {\kern 1pt} {\kern 1pt} {\kern 1pt} {\kern 1pt} {\kern 1pt} {\kern 1pt} s.t.{\kern 1pt} {\kern 1pt} {\kern 1pt} {\kern 1pt} F = L + S \\
\end{split}
\end{equation}
where $L$ is a low-rank matrix, indicating the background, i.e., repeated patterns. $S$ is a sparse matrix, and it represents the defective objects.

Since the above problem is NP-hard, it is difficult to approximate the optimal solution of (7), so the convex surrogate is adopted as follows:

\begin{equation}
\left( {L^* ,S^* } \right) = \arg {\kern 1pt} {\kern 1pt} {\kern 1pt} \mathop {\min }\limits_{\left( {L,S} \right)} {\kern 1pt} {\kern 1pt} {\kern 1pt} \left( {\left\| L \right\|_*  + \lambda \left\| S \right\|_1 } \right){\kern 1pt} {\kern 1pt} {\kern 1pt} {\kern 1pt} {\kern 1pt} {\kern 1pt} {\kern 1pt} s.t.{\kern 1pt} {\kern 1pt} {\kern 1pt} {\kern 1pt} {\kern 1pt} {\kern 1pt}  = L + S
\end{equation}
where $\left\| L \right\|_*$ is the nuclear norm of $L$, $\left\| {{\kern 1pt} {\kern 1pt} {\kern 1pt} .{\kern 1pt} {\kern 1pt} {\kern 1pt} {\kern 1pt} } \right\|_1$ indicates the $l_1$ norm, $\lambda$ is a weighted factor that controls the low-rank and sparsity degree.

Using the nuclear norm as a convex surrogate in the first term of Eq. (8), i.e., rank minimization problem is correct, while non-convex optimization toward the rank minimization problem could lead to better recovery results.

In this paper, a smooth but non-convex surrogate of the rank is adopted instead of the nuclear norm. For a given matrix with a symmetric positive semi-definite $X \in R^{n \times n}$, the rank minimization problem can be approximately surrogate by minimizing the following equation [34]:
\begin{equation}
E(X,\xi ) = \log \det (X + \xi I)
\end{equation}
where $\xi$ is a positive scalar. $E\left( {X,\xi } \right)$ approximates the sum of the logarithm of singular values, thus it is smooth and non-convex. The log det as a non-convex surrogate of the rank has also been more carefully proofed from an information-theoretic perspective. As shown in Figure 3, the surrogate function $E\left( {X,\xi } \right)$ can better approximate the rank than the nuclear norm.

\begin{figure}[ht]
  \centering
  \includegraphics[width=0.4\textwidth]{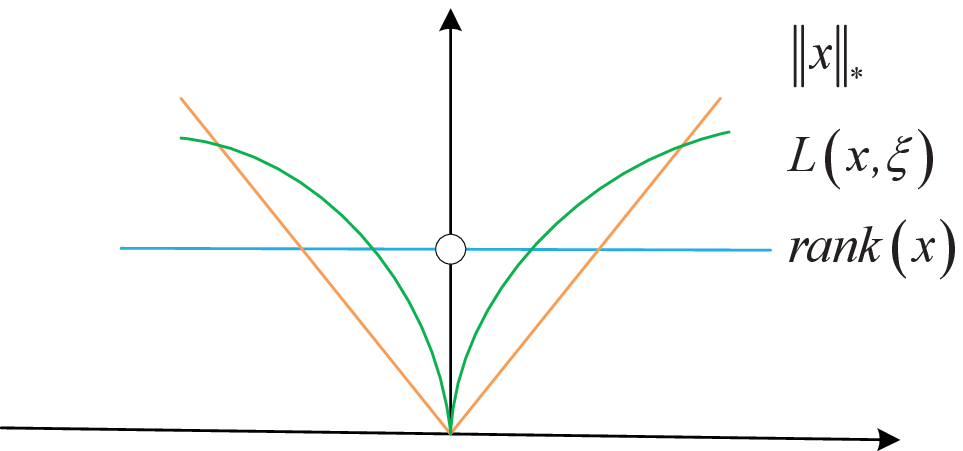}\\
  \caption{Comparison of $L\left( {x,\xi } \right)$, $rank\left( x \right)$, $\left\| x \right\|_*$  in the case of a scalar.}
\end{figure}

For the low-rank matrix $L$, Eq.(9) is rewritten as follows:
\begin{equation}
\begin{split}
 L(L,\xi ) = \log \det ((LL^T )^{1/2}  + \xi I) \\
 {\kern 1pt} {\kern 1pt} {\kern 1pt} {\kern 1pt} {\kern 1pt} {\kern 1pt} {\kern 1pt} {\kern 1pt} {\kern 1pt} {\kern 1pt} {\kern 1pt} {\kern 1pt} {\kern 1pt} {\kern 1pt} {\kern 1pt} {\kern 1pt} {\kern 1pt} {\kern 1pt} {\kern 1pt} {\kern 1pt} {\kern 1pt} {\kern 1pt} {\kern 1pt} {\kern 1pt} {\kern 1pt} {\kern 1pt} {\kern 1pt} {\kern 1pt} {\kern 1pt} {\kern 1pt}  = \log \det (U\Sigma ^{1/2} U^{ - 1}  + \xi I) \\
 {\kern 1pt} {\kern 1pt} {\kern 1pt} {\kern 1pt} {\kern 1pt} {\kern 1pt} {\kern 1pt} {\kern 1pt} {\kern 1pt} {\kern 1pt} {\kern 1pt} {\kern 1pt} {\kern 1pt} {\kern 1pt} {\kern 1pt} {\kern 1pt} {\kern 1pt} {\kern 1pt} {\kern 1pt} {\kern 1pt} {\kern 1pt} {\kern 1pt} {\kern 1pt} {\kern 1pt} {\kern 1pt} {\kern 1pt} {\kern 1pt} {\kern 1pt} {\kern 1pt} {\kern 1pt}  = \log \det (\Sigma ^{1/2}  + \xi I) \\
\end{split}
\end{equation}
where $\Sigma$ is the diagonal matrix whose diagonal elements are eigenvalues of matrix $LL^T$, i.e., $LL^T  = U\Sigma U^{ - 1}$, meanwhile, $\Sigma ^{1/2}$ is also the diagonal matrix whose diagonal elements are the singular values of the matrix $L$. Hence, $L\left( {L,\xi } \right)$ is a log det(¡¤) surrogate function of $rank\left( L \right)$ obtained by setting $X = \left( {LL^T } \right)^{1/2}$. Finally, Eq.(8) can be rewritten as follows:
\begin{equation}
(L^* ,S^* ) = \arg {\kern 1pt} {\kern 1pt} \mathop {\min }\limits_{(L,S)} {\kern 1pt} {\kern 1pt} (L(L,\xi ) + \lambda \left\| S \right\|_1 ){\kern 1pt} {\kern 1pt} {\kern 1pt} {\kern 1pt} s.t.{\kern 1pt} {\kern 1pt} {\kern 1pt} {\kern 1pt} {\kern 1pt} F = L + S
\end{equation}

\subsection{Optimal solution of the model}
The alternative direction method of multipliers (ADMM) demonstrates a good balance between efficiency and accuracy in solving optimization problems. In this paper, ADMM has been adopted to solve Eq. (11).

The Augmented Lagrange Multiplier function of Eq.(11) is as follows:
\begin{equation}
\begin{split}
 L(L,S,Z) = L(L,\xi ) + \lambda \left\| S \right\|_1  + \frac{\beta }{2}\left\| {L + S - F} \right\|_F^2  \\
 {\kern 1pt} {\kern 1pt} {\kern 1pt} {\kern 1pt} {\kern 1pt} {\kern 1pt} {\kern 1pt} {\kern 1pt} {\kern 1pt} {\kern 1pt} {\kern 1pt} {\kern 1pt} {\kern 1pt} {\kern 1pt} {\kern 1pt} {\kern 1pt} {\kern 1pt} {\kern 1pt} {\kern 1pt} {\kern 1pt} {\kern 1pt} {\kern 1pt} {\kern 1pt} {\kern 1pt} {\kern 1pt} {\kern 1pt} {\kern 1pt} {\kern 1pt} {\kern 1pt} {\kern 1pt} {\kern 1pt} {\kern 1pt} {\kern 1pt} {\kern 1pt} {\kern 1pt} {\kern 1pt} {\kern 1pt} {\kern 1pt} {\kern 1pt} {\kern 1pt} {\kern 1pt} {\kern 1pt} {\kern 1pt} {\kern 1pt} {\kern 1pt} {\kern 1pt} {\kern 1pt} {\kern 1pt} {\kern 1pt} {\kern 1pt} {\kern 1pt} {\kern 1pt} {\kern 1pt} {\kern 1pt} {\kern 1pt} {\kern 1pt} {\kern 1pt}  - \left\langle {Z,L + S - F} \right\rangle  \\
\end{split}
\end{equation}
where $Z \in R^{m \times n}$ is the multiplier of the linear constraint, $\beta  > 0$ is the penalty parameter for the violation of the linear constraint, $\left\langle  \cdot  \right\rangle$   is the inner product and $\left\|  \cdot  \right\|_F$ is the induced Frobenius norm. The proposed objective function can be solved by alternatively minimizing the objective function with respect to the $L$, $S$ and the multiplier $Z$. It can be described as solving the following three sub-problems:
\begin{equation}
\left\{ \begin{array}{l}
 L_{k + 1}  = \arg \min _L L(L,S_k ,Z_k ;\beta ) \\
 S_{k + 1}  = \arg \min _S L(L_{k + 1} ,S,Z_k ;\beta ) \\
 Z_{k + 1}  = Z_k  - \beta \left( {L_{k + 1}  + S_{k + 1}  - F} \right) \\
 \end{array} \right.
\end{equation}

For the first sub-problem in (13) which solves for $L$ at fixed $S$ and $Z$, it can be explicitly represented as the following form:
\begin{equation}
\begin{split}
 L^*  = \mathop {\arg \min }\limits_L \sum\nolimits_{j = 1}^{n_0 } {\log (\sigma _j \left( L \right) + \xi )}  + \frac{\beta }{2}\left\| {L + S - F} \right\|_F^2  \\
 {\kern 1pt} {\kern 1pt} {\kern 1pt} {\kern 1pt} {\kern 1pt} {\kern 1pt} {\kern 1pt} {\kern 1pt} {\kern 1pt} {\kern 1pt} {\kern 1pt} {\kern 1pt} {\kern 1pt} {\kern 1pt}  - \left\langle {Z,L + S - F} \right\rangle  \\
\end{split}
\end{equation}
where $n_0  = \min \left\{ {m,n} \right\}$, and $\sigma _j \left( L \right)$  indicates the $j$-th singular value of $L$. For simplicity, we use $\sigma _j$ to denote $\sigma _j \left( L \right)$ . Even though $\sum\nolimits_{j = 1}^n {{\rm{log}}\left( {\sigma _j  + \xi } \right)}$ is non-convex, it can be solved by utilizing a local minimization approach. We define the equality $
f\left( \sigma  \right) = \sum\nolimits_{j = 1}^n {{\rm{log}}\left( {\sigma _j  + \xi } \right)}$. Then $f\left( \sigma  \right)$ can be approximated by using its first-order Taylor expansion, as follows:
\begin{equation}
f\left( \sigma  \right) = f\left( {\sigma ^{(k)} } \right) + \left\langle {\nabla f\left( {\sigma ^{(k)} } \right),\sigma  - \sigma ^{(k)} } \right\rangle
\end{equation}
where $\sigma ^{(k)}$ is the solution obtained in the $k$-th iteration. Therefore, Eq.(14) can be solved by iteratively solving:
\begin{equation}
\begin{split}
 L^{(k + 1)}  = \mathop {\arg \min }\limits_L \frac{\beta }{2}\left\| {L + S - F - \frac{Z}{\beta }} \right\|_F^2  \\
 {\kern 1pt} {\kern 1pt}  + \sum\nolimits_{j = 1}^n {\frac{{\sigma _j }}{{\sigma _j^{(k)}  + \xi }}}  \\
\end{split}
\end{equation}
where we use the fact that $\nabla f(\sigma ^{(k)} ) = \sum\nolimits_{j = 1}^{n_0 } {\frac{1}{{\sigma _j^{(k)}  + \xi }}}$ and ignore the constants in Eq.(14). For convenience, we rewrite Eq.(16) as:
\begin{equation}
L^{(k + 1)}  = \mathop {\arg \min }\limits_L \frac{1}{2}\left\| {L + S - F - \frac{Z}{\beta }} \right\|_F^2  + \tau \varphi \left( {L,\omega ^{(k)} } \right)
\end{equation}
where $\tau  = 1/\beta$. $\varphi (L,\omega ) = \sum\nolimits_j^{n_0 } {\omega _j^{(k)} } \sigma _j$ indicates a weighted nuclear norm whose weights $\omega _j^{(k)}  = 1/\left( {\sigma _j^{(k)}  + \xi } \right)$. Note that the weights are ascending, since the singular values $\sigma _j$ are ordered in a descending order.

In general, for a real matrix, the weighted nuclear norm is a convex function only if the weights are descending, and the optimal solution to Eq. (17) is obtained by a weighted singular value thresholding operator, referred to as the proximal operator. In this paper, the weights are ascending, thus Eq. (17) is non-convex. It is therefore difficult to find its global minimizer. Nevertheless, we could find that the weighted singular value thresholding gives one minimizer to Eq. (17) via Theorem 1 (Proximal Operator of Weighted Nuclear Norm) [35]. According to this Theorem, we can obtain the low-rank matrix at the $(k+1)$-th iteration by
\begin{equation}
L^{\left( {k + 1} \right)}  = U\left( {\Sigma  - \tau diag\left( {\omega ^{\left( k \right)} } \right)} \right)_ +  V^T
\end{equation}
where $U\Sigma V^T$ is the SVD of the feature matrix $F$, and $\omega _j^{(k)}  = 1/\left( {\sigma _j^{(k)}  + \xi } \right)$. Even though the weighted thresholding is only a local minimizer, it always leads to a decrease in the objective function value. In this paper, the initial value $\omega ^{\left( 0 \right)} {\kern 1pt} {\kern 1pt} {\kern 1pt}$ is set to $\left[ {1,1,...1} \right]^T$.

After solving the low-rank matrix L, the sparse matrix S can be solved by fixing L and Z. Indeed, we can easily obtain the solution using the widely-used shrinkage problem [36]:
\begin{equation}
S^{k + 1}  = \frac{1}{\beta }Z^k  - L^k  + F - P_{\Omega _\infty ^{\gamma /\beta } } \left[ {\frac{1}{\beta }Z^k  - L^k  + F} \right]
\end{equation}
where $P_{\Omega _\infty ^{\gamma /\beta } }$ indicts the Euclidean projection onto:
\begin{equation}
\Omega _\infty ^{\gamma /\beta } : = \left\{ {X \in R^{n \times n} \left| { - \gamma /\beta  \le X_{ij}  \le \gamma /\beta } \right.} \right\}
\end{equation}
Then the multipliers $Z$ can be updated as follows:
\begin{equation}
Z^{k + 1}  = Z^k  - \beta (L^{k + 1}  + S^{k + 1}  - F)
\end{equation}

\subsection{The generation and segmentation of the saliency map}
 \textbf{Saliency map generation.} After decomposing the feature matrix $F$ of the given patterned fabric image into the low-rank matrix $L$, which corresponds to the patterned fabric background, and a sparse matrix $S$, which corresponds to the defective regions by the aforementioned methods. A saliency map $M$ is generated by the $l_1$-norm of each column $S_i$ in $S$, and it is described as follows:
\begin{equation}
M(I_i ) = \left\| {S_i } \right\|_1
\end{equation}

The larger value of $M\left( {I_i } \right)$ indicates that the block $I_i$ has higher probability of being defective.

\textbf{The segmentation of saliency map.}

1). Denoise the saliency map $M$ to generate a new saliency map $\mathop M\limits^ \wedge$£º
\begin{equation}
\mathop M\limits^ \wedge   = g*(M \circ M)
\end{equation}
where $g$ is the radius of the circular smoothing filter and "$\circ$" denotes the Hadamard inner product operator and "*" is the convolution operator.

2). Transform the saliency map $\mathop M\limits^ \wedge$ into a gray-scale image $G$ £º
\begin{equation}
G = \frac{{\mathop M\limits^ \wedge   - \min (\mathop M\limits^ \wedge  )}}{{\max (\mathop M\limits^ \wedge  ) - \min (\mathop M\limits^ \wedge  )}} \times 255
\end{equation}

3). Segment $G$ by using improved adaptive threshold algorithm [37] to locate the defect regions.

\section{Experimental results and analysis}
In order to verify the effectiveness of the proposed algorithm, we chose images from the dot-, box-, and star-patterned fabric databases for performance evaluation. The size of the fabric image is 512 pixels $\times$ 512 pixels. The dot-patterned fabric database contains 30 defect-free and 30 defective images, the star-patterned fabric database contains 25 defect-free and 25 defective images, and the box-patterned fabric database contains 30 defect-free and 26 defective images. All the defective images have a corresponding ground-truth that shows the defect-free regions as black and defective regions as white. All experiments in this paper were implemented in an Inter(R) Core(TM) i3-2120 3.3GHZCPU environment, using software MATLAB 2011a.

\emph{A. Qualitative Results}

The test patterned fabric image matrix was divided into the superposition of two parts, i.e., a low-rank matrix and a sparse matrix, and the visual saliency map was generated from the sparse matrix and outstands the defective regions. Thus a simple threshold method can easily localize the defect region.

The feature extraction method and detection model are equally important for fabric defect detection. In order to validate the effectiveness of our method, we firstly compared the saliency maps generated by different descriptors, including Gabor [32], HOG [33], and different detection models, such as the template matching model (TMM) [38] and the context analysis model (CAM) [39] with our method, as shown in Figure 4. The first column consists of the original images, the second column consists of saliency maps generated by the Gabor feature [32] and the LR detection model, the third column consists of the saliency maps generated by the HOG feature [33] and the LR detection model, the fourth column consists of the saliency maps generated by the GHOG feature and the TMM [38] detection model, the fifth column consists of the saliency maps generated by the GHOG feature and the CAM [39] detection model, and the last column consists of the saliency maps generated by our method (GHOG feature and the LR detection model). From the second column and the third column in Figure 4, we can conclude that the detection result based on the Gabor or HOG features cannot outstand the defect region for star-patterned fabric and box-patterned fabric images, but they can outstand the detect region for dot-patterned fabric images. On the other hand, from the fourth column and the fifth column in Figure 4, the saliency map generated by the GHOG feature and the TMM or CAM detection models cannot outstand the defect region. In the last column we can see that the saliency map generated by our method can efficiently outstand the defect region for all the three types of images. And the performance of these methods can be concluded as Table1.
\begin{table*}[htbp]
\centering
 \caption{\label{tab:test}Comparisons with other methods}
 \resizebox{\textwidth}{!}{
\begin{tabular}{lp{2cm}p{2.5cm}p{2.5cm}p{2.5cm}p{2.5cm}}
  \toprule
  Methods$/$ fabric image & Gabor+LR & HOG+LR & GHOG+TMM & GHOG+CAM & GHOG+LR  \\
  star-patterned & $\times$ & $\times$ & $\times$ & $\times$ & $\surd$ \\
  box-patterned & $\times$ & $\times$ & $\times$ & $\times$ &$\surd$ \\
  dot-patterned & $\surd$ & $\surd$ & $\times$ & $\times$ &$\surd$  \\
  \bottomrule
 \end{tabular}}
\end{table*}

\begin{figure*}[htp]
\centering
\subfigure[Saliency map for star-patterned fabric image]{\includegraphics[width= 0.495\textwidth]{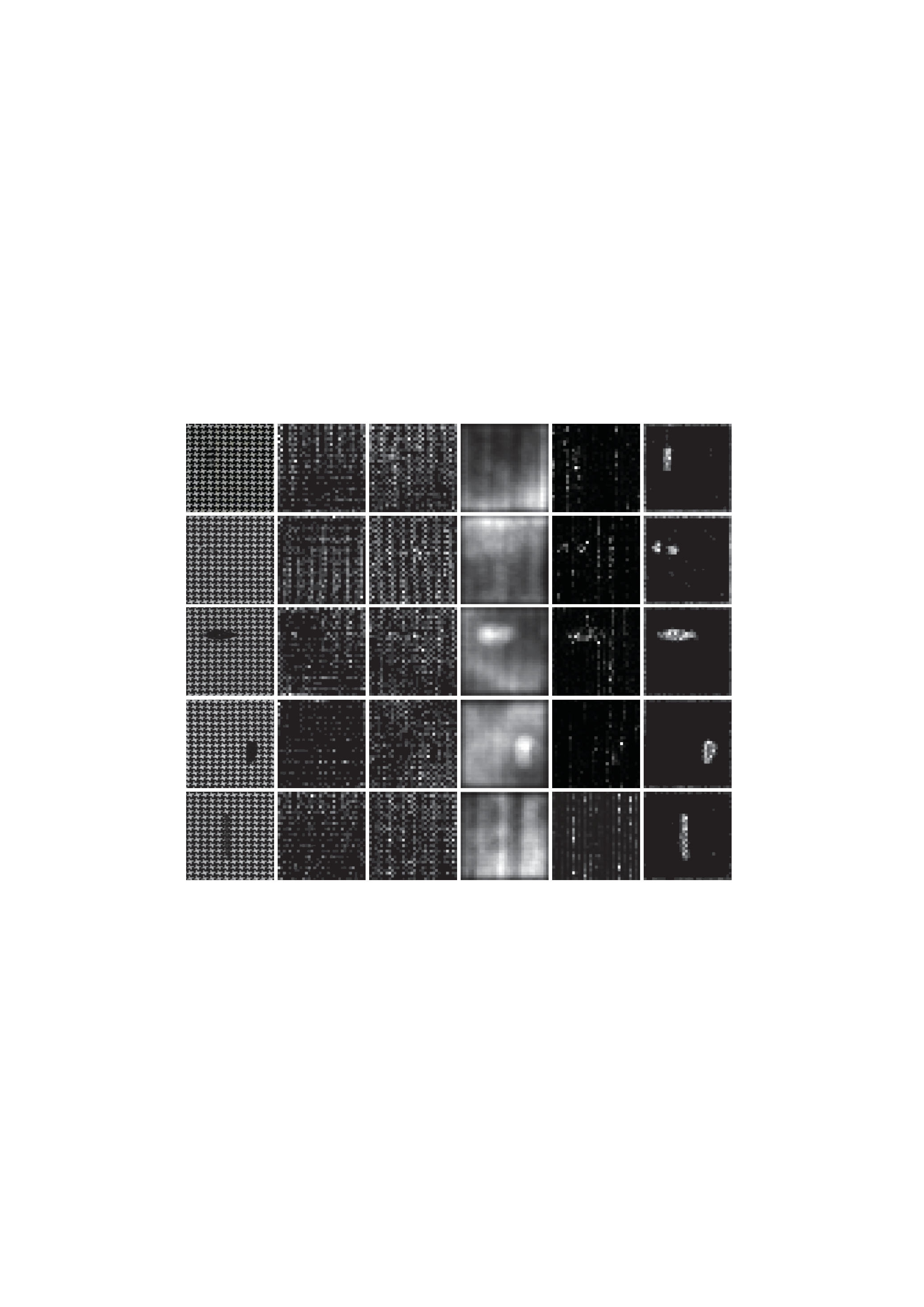}}\quad
\subfigure[Saliency map for box-patterned fabric imagec]{\includegraphics[width= 0.495\textwidth]{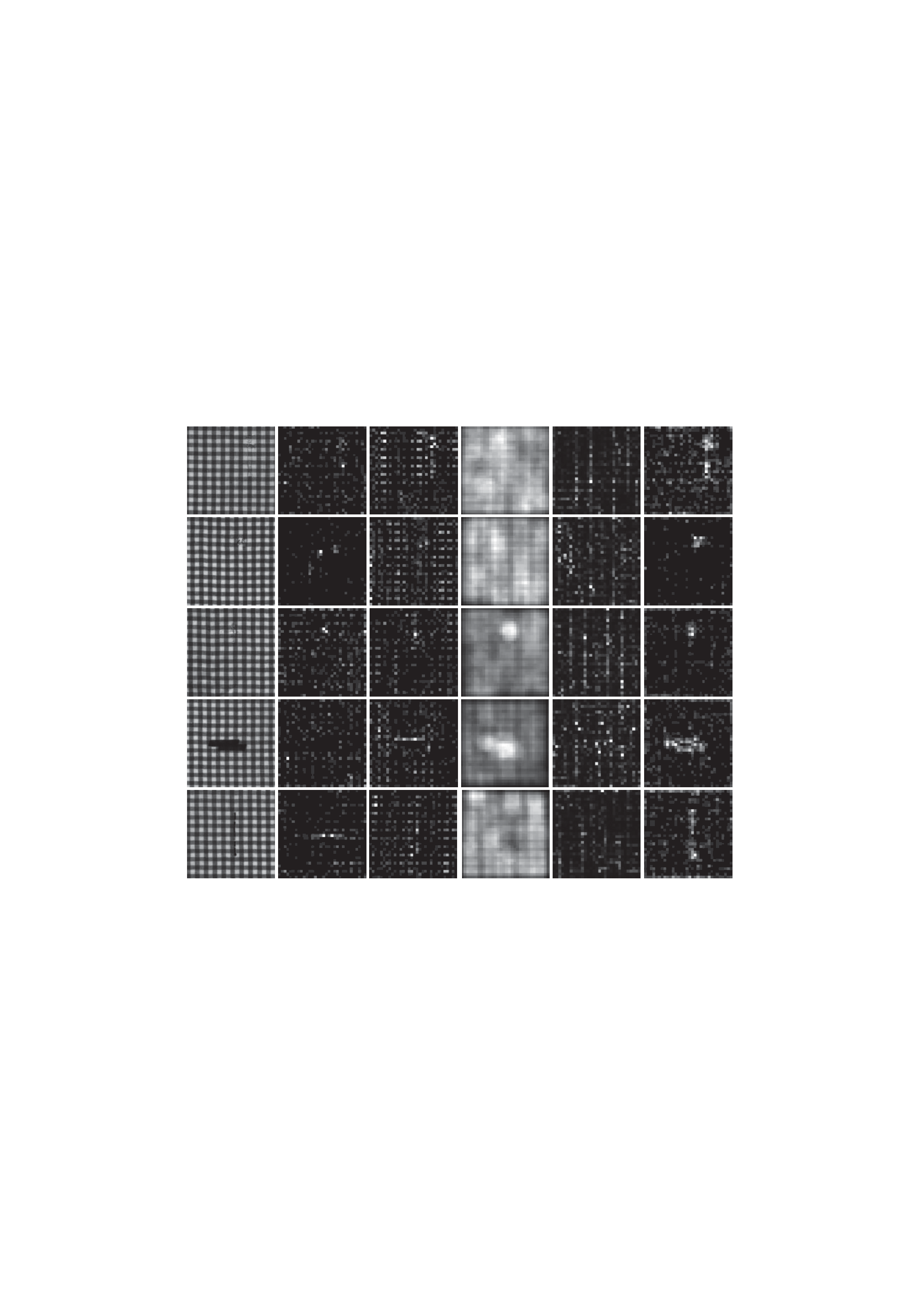}}\quad
\subfigure[Saliency map for dot-patterned fabric image]{\includegraphics[width= 0.495\textwidth]{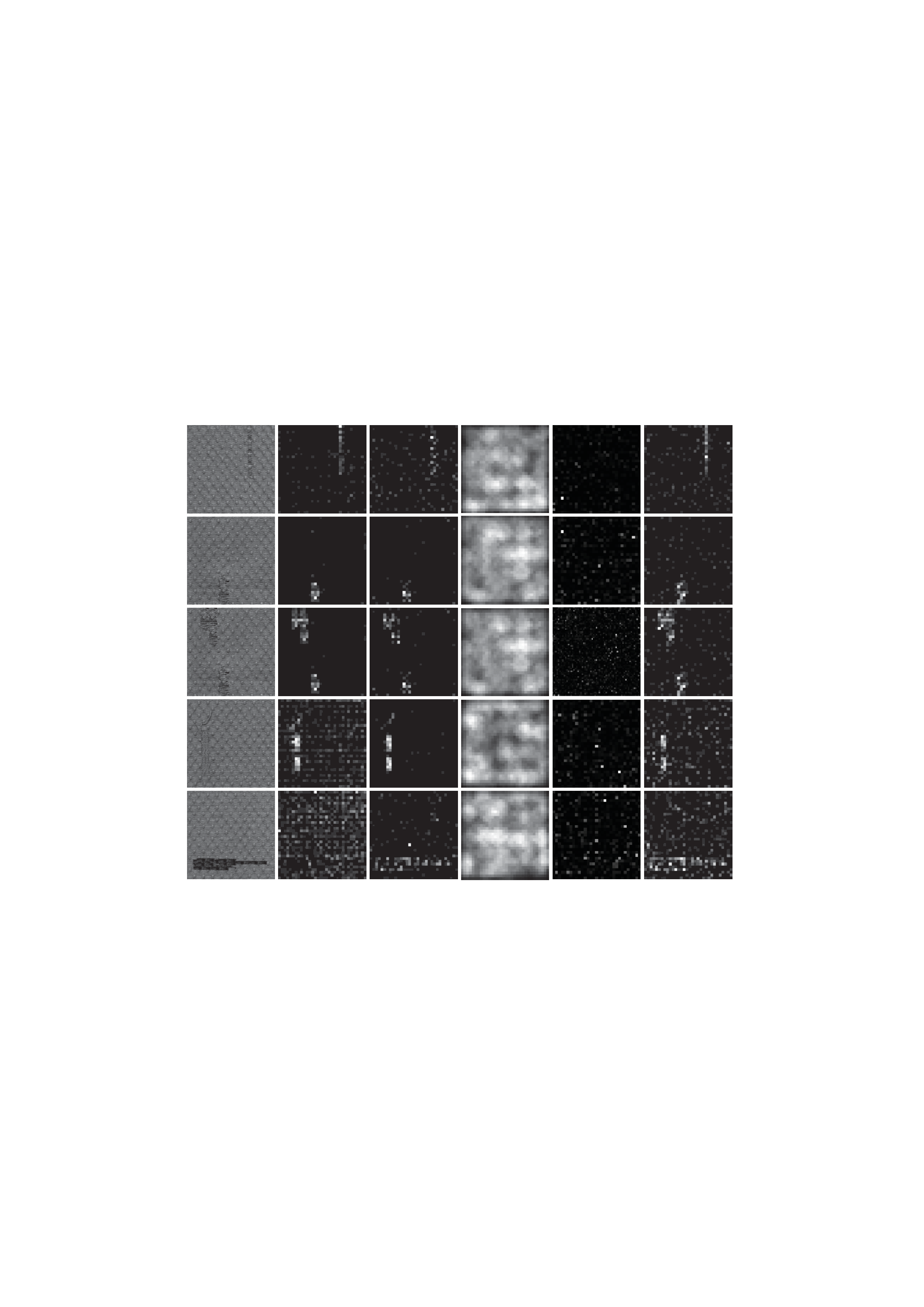}}\quad
  \caption{The saliency maps generated by different features and detection model. The first column shows the original images, the second column shows saliency maps generated by Gabor [32] with the LR model (Gabor + LR), the third column shows the saliency maps generated by HOG [33] with LR model (HOG + LR), the fourth column shows the saliency maps generated by TDVSM [38](GHOG + TDVSM), the fifth column shows the saliency maps generated by LSF-GSA [39](GHOG + LSF-GSA), and the last column shows the saliency maps generated by our method(GHOG + LR).}
\end{figure*}

From Figure 4 and Table 1, we can see that the proposed defection method based on GHOG and LR is suitable for pattern fabric defect detection.

In addition, we compared our method with some state of the art visual saliency models, including the wavelet transform (WT) method [40], the prior guided least squares regression method (PGLSR) [41], the textural differential visual saliency model (TDVSM) [38] and the local statistic features and global saliency analysis model (LSF-GSA) [39].

As is shown in Figure 5, WT [40] first transformed the image into a frequency domain, then generated the saliency map by analyzing the wavelet coefficient. However, even in a normal background with a complicated pattern, its wavelet coefficients are larger, which will lead to incorrect detection results. The PGLSR method [41] could effectively detect defects in the patterned fabric, but similarities in texture between the background and the defect lead to inaccurate shape descriptions of the defects. In Li et al. [38], saliency was calculated by comparing their textural features with the average texture features, it  obtains a successful performance for fabric images with a stochastic texture, while the method fails to detect defects in a patterned fabric image, especially in box-patterned fabrics, as shown in Figure.5(b). Liu et al. [39] detected fabric defects by using the local statistic feature and global saliency analysis, while the detection results have a large amount of noise. Our method generates a saliency map by combining the GHOG feature descriptor with a low-rank decomposition model, and it effectively highlights the defect regions. Subsequently, the detection results are obtained by segmenting the generated saliency map.
\begin{figure*}[htp]
\centering
\subfigure[Saliency map for star-patterned fabric image]{\includegraphics[width= 0.495\textwidth]{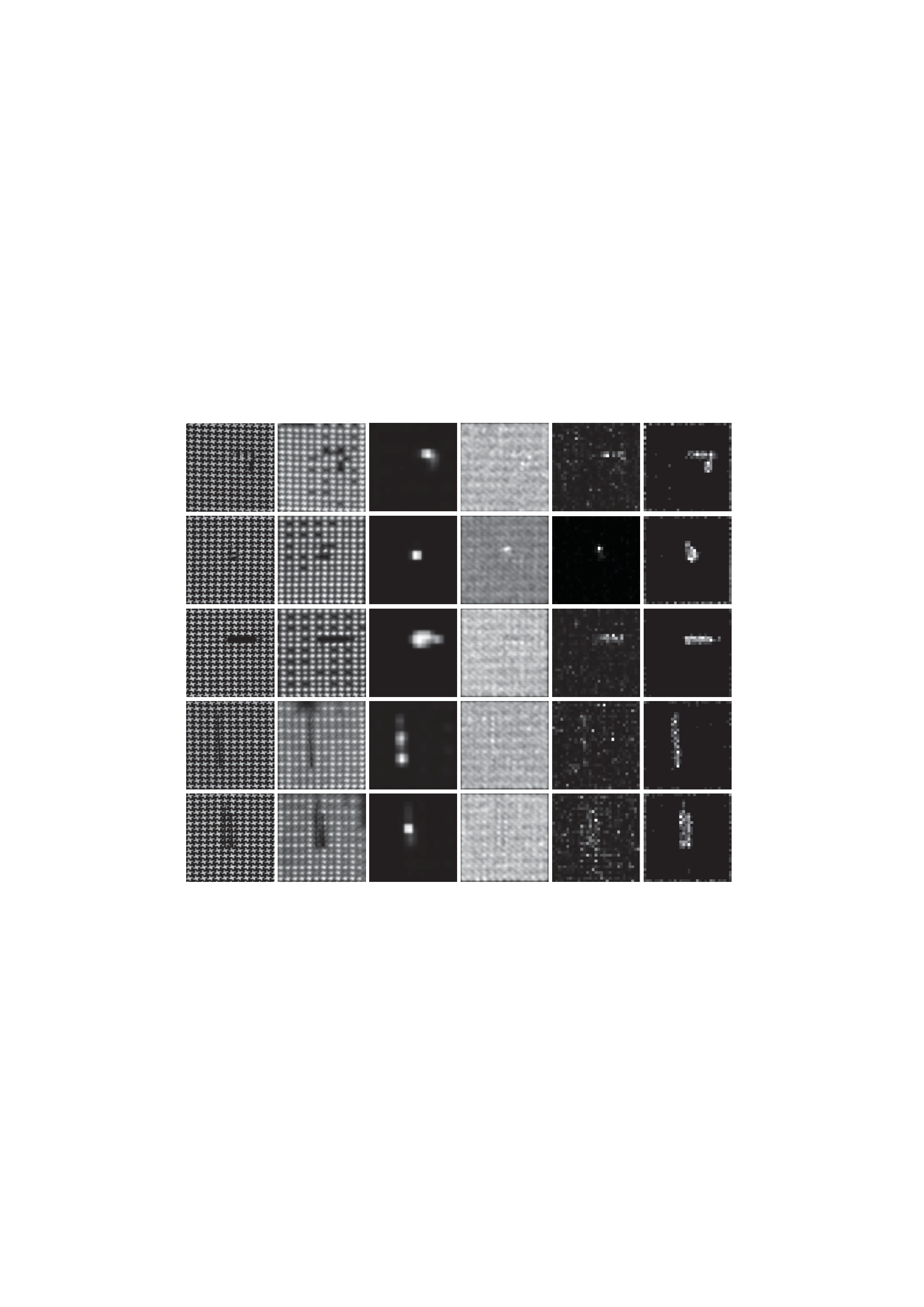}}\quad
\subfigure[Saliency map for box-patterned fabric image]{\includegraphics[width= 0.495\textwidth]{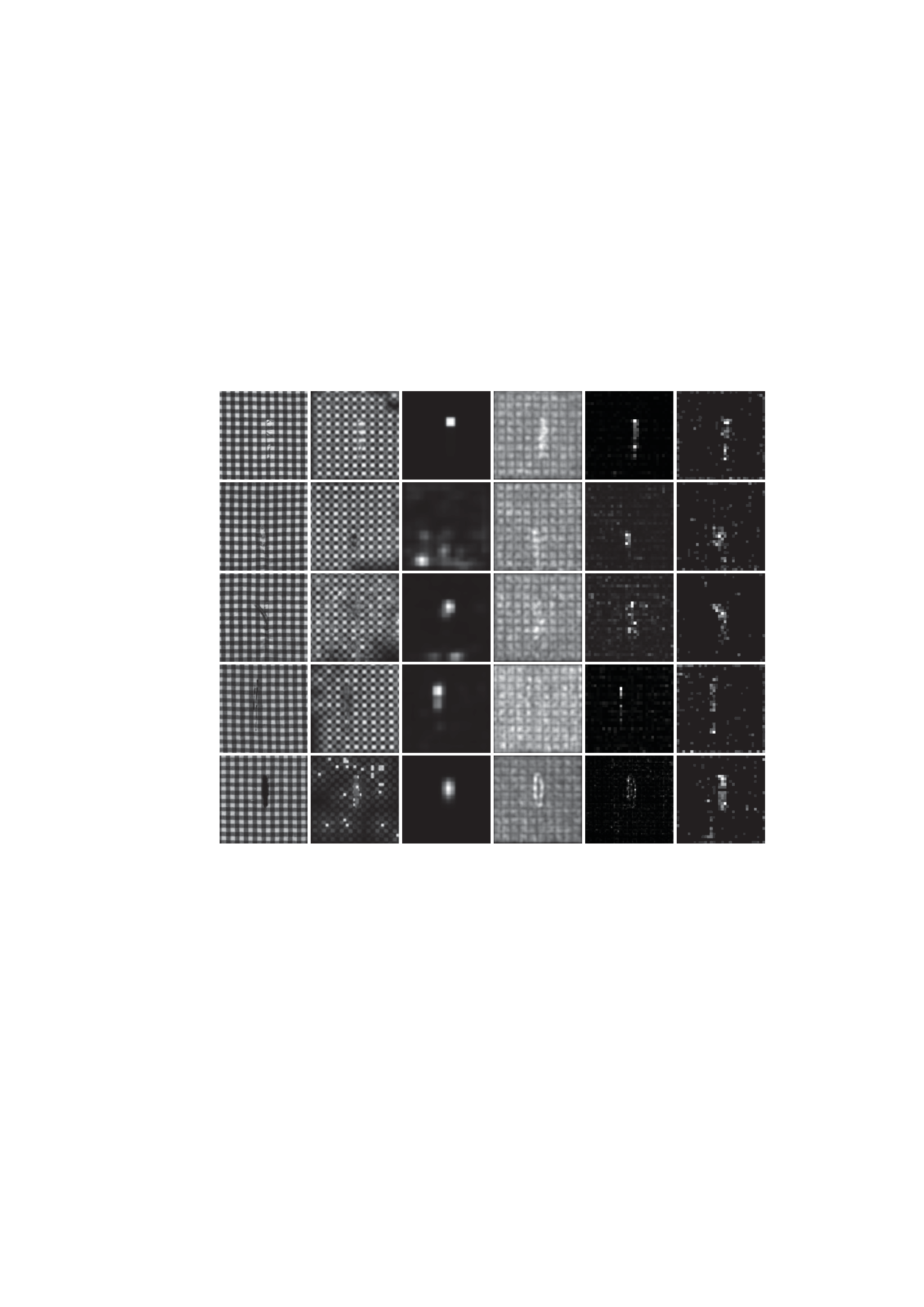}}\quad
\subfigure[Saliency map for dot-patterned fabric image]{\includegraphics[width= 0.495\textwidth]{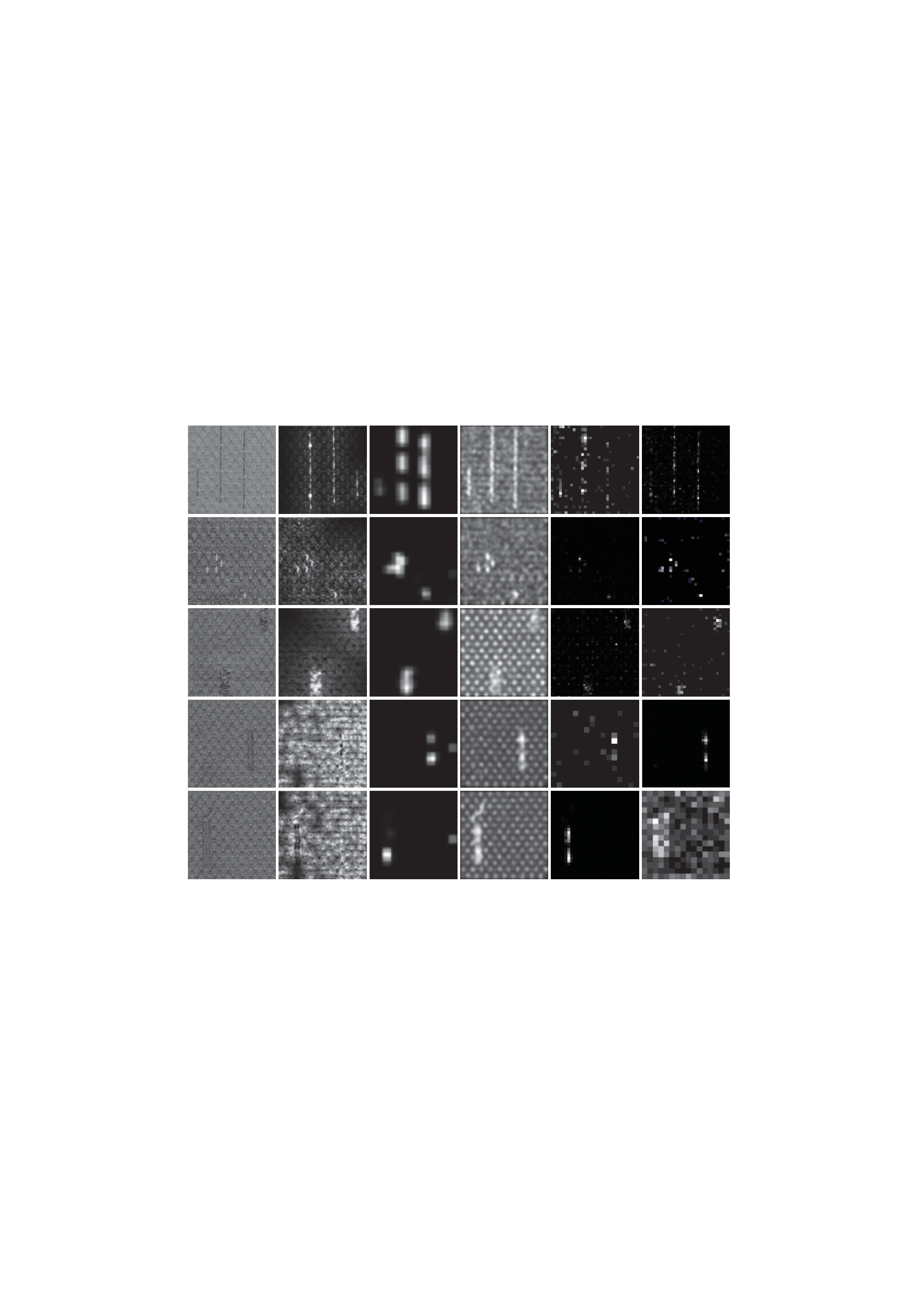}}\quad
  \caption{The saliency maps of our method compared with other state-of-the-art saliency models. The first column shows the original images, the second column shows saliency maps generated by WT [40], the third column shows the saliency maps generated by PGLSR [41], the fourth column shows the saliency maps generated by TDVSM [38], the sixth column shows the saliency maps generated by LSF-GSA [39], and the last column shows the saliency maps generated by our method.}
\end{figure*}
In order to further verify the effectiveness of our proposed method, we compared our detection method with the existing fabric detection algorithm. Because most of the traditional detection methods are invalid for detecting patterned fabric defect, in this paper, we only compared the proposed method with the other two valid detection methods for detecting patterned fabrics. The experimental results are described in Figure 6, where the white pixels represent the defect regions, and the black pixels represent the background. The first row represents the detection results generated by PGLSR [41], the second row represents the detection results generated by LSF-GSA [39], the third row indicates the detection results generated by our method and the last row indicates the ground-truth images. In Figure 6, we can see that the other two methods (as shown in the first two rows) can almost localize the defect region, but the detected shape of the defect is different from the ground truth. Our detection results (as shown in the third row) are similar to the ground truth (as shown in the last row), can efficiently localize the defect regions, and the detected shape of the defect is similar to the ground truth. This demonstrates our proposed method is superior to the other two methods.
\begin{figure*}[htp]
\centering
\subfigure[Detection result for star-patterned fabric image]{\includegraphics[width= 0.495\textwidth]{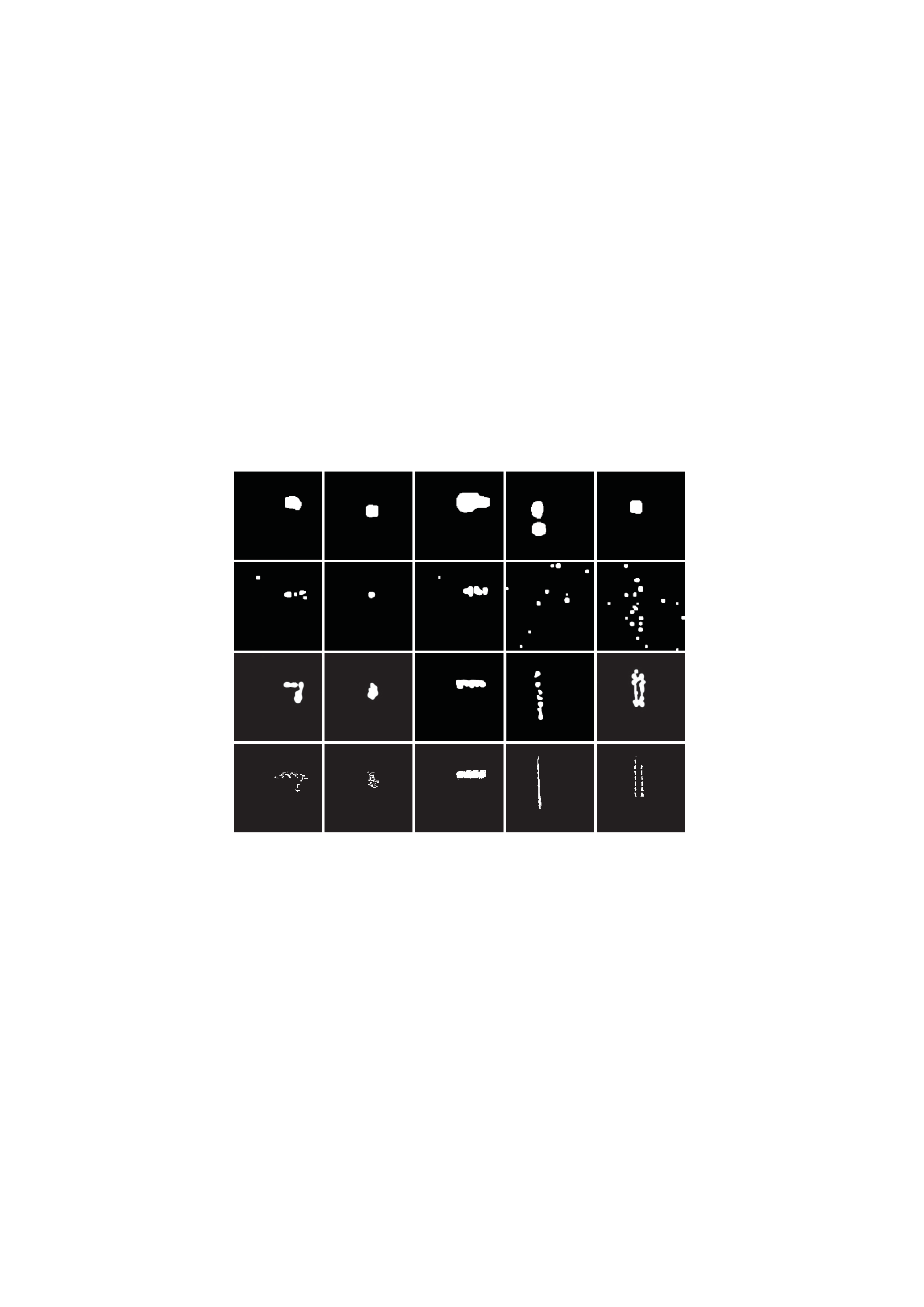}}\quad
\subfigure[Detection result for box-patterned fabric image]{\includegraphics[width= 0.495\textwidth]{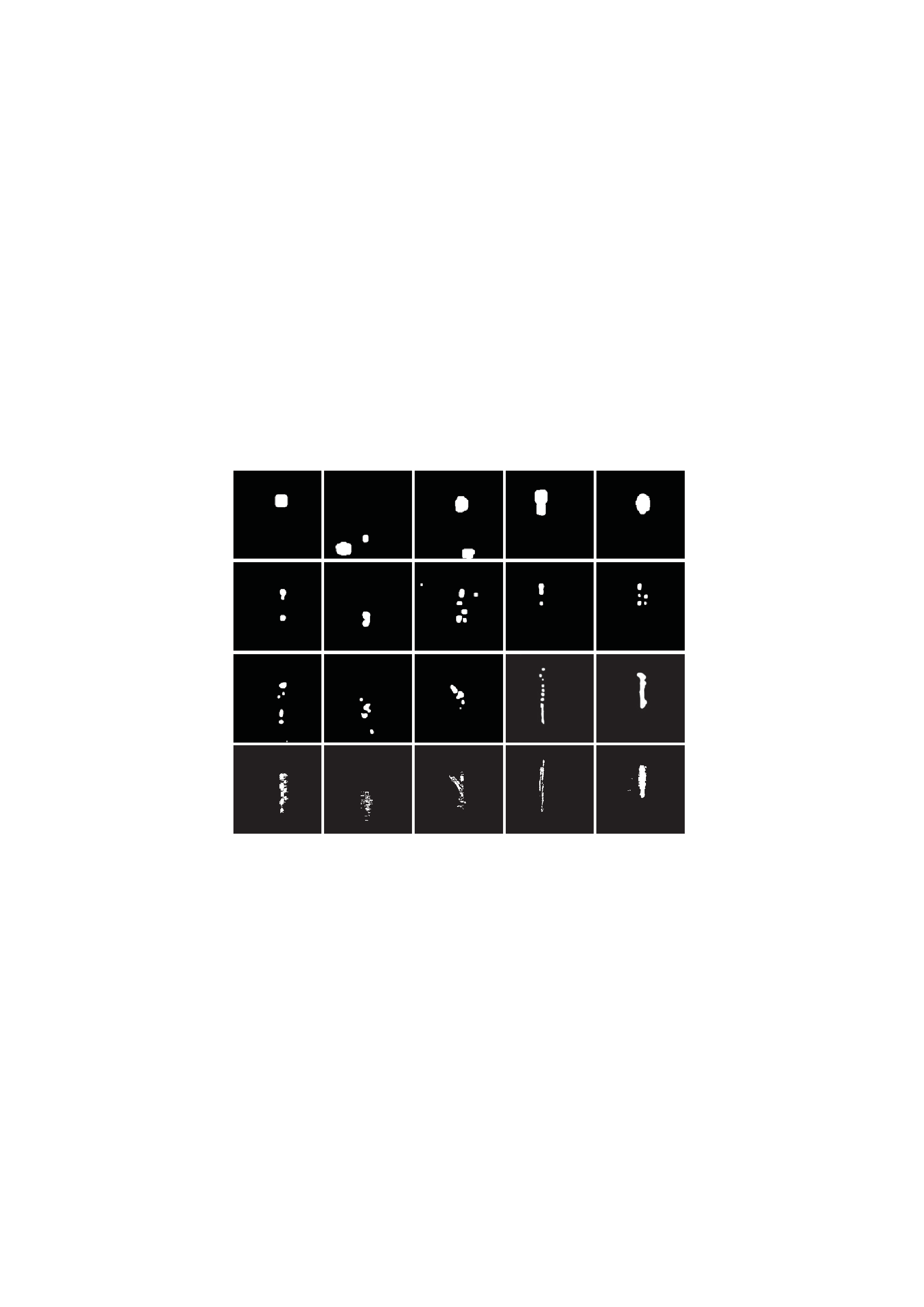}}\quad
\subfigure[Detection result for dot-patterned fabric image]{\includegraphics[width= 0.495\textwidth]{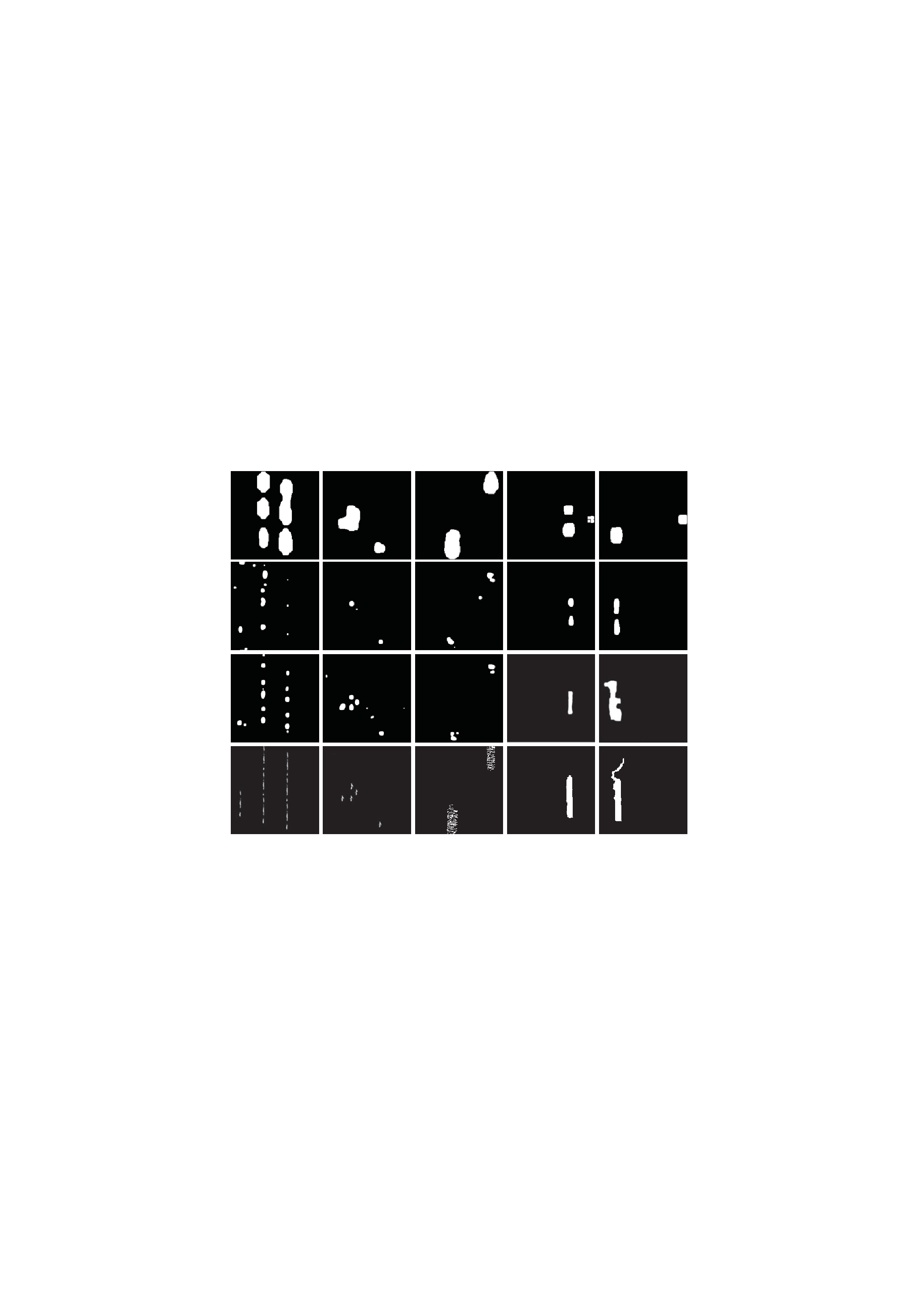}}\quad
  \caption{Detection results of our method compared with other fabric defect detection methods. The first row shows the detection results of PGLSR method, the second row shows the detection results of LSF-GSA method, the third row shows the detection results of our method, the last row shows the ground-truth images.}
\end{figure*}

\emph{B. Quantitative Evaluations}

In order to further evaluate the performance of our proposed method, the receiver operating characteristic curve (ROC) is adopted, as shown in Figure 7. Figure.7(a) shows the ROC curves of star-patterned fabric images, Figure.7(b) shows the ROC curves of box-patterned fabric images, and Figure.7(c) shows the ROC curves of dot-patterned fabric images. From Figure 7(a) and (b), we can see that our method outperforms the other three methods for star-patterned and box-patterned fabrics. This demonstrates the effectiveness of our proposed method. However, our proposed method is unsatisfactory regarding dot-patterned fabrics, as shown in Figure.7(c). The proposed GHOG feature is orientation-aware, while the orientation features of the dot-patterned fabric are indistinct; therefore, GHOG is not enough to characterize the dot-patterned fabric. In fact, for most of the fabric samples, they are weaved using warp and weft, and have distinct directions.

\begin{figure*}[htp]
\centering
\subfigure[Star-patterned ROC curves]{\includegraphics[width= 0.32\textwidth]{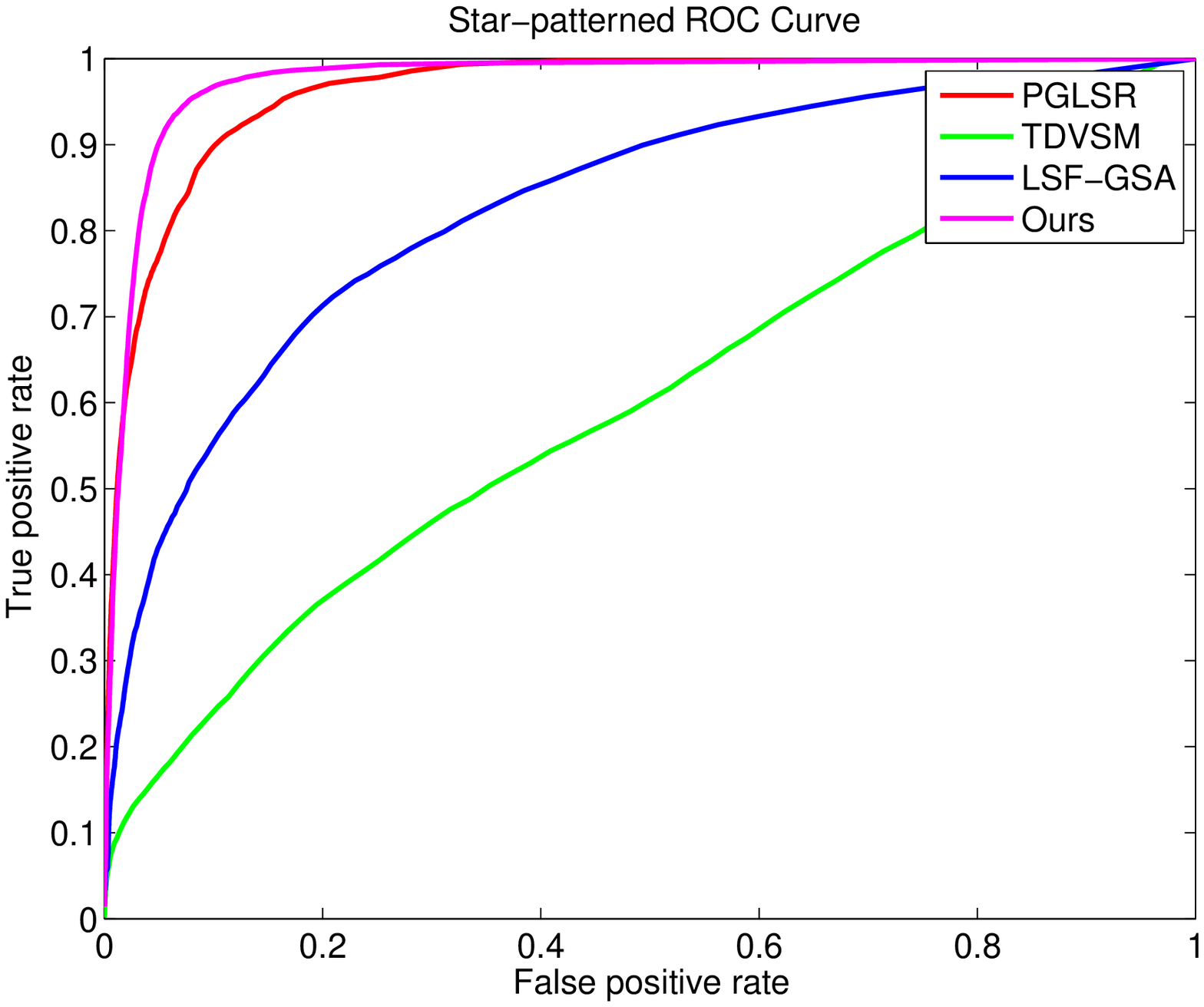}}\quad
\subfigure[Box-patterned ROC curves]{\includegraphics[width= 0.32\textwidth]{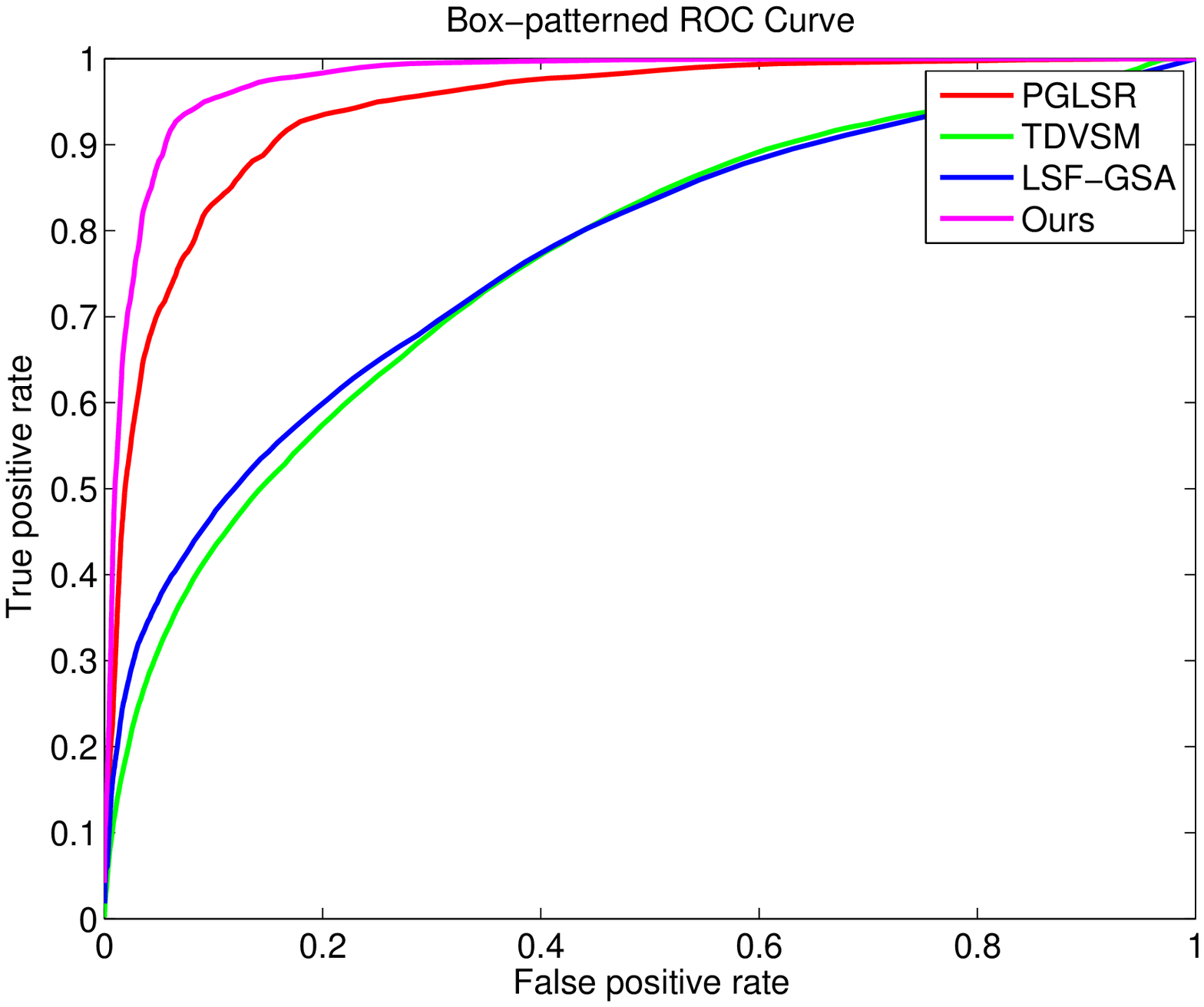}}\quad
\subfigure[Dot-patterned ROC curves]{\includegraphics[width= 0.32\textwidth]{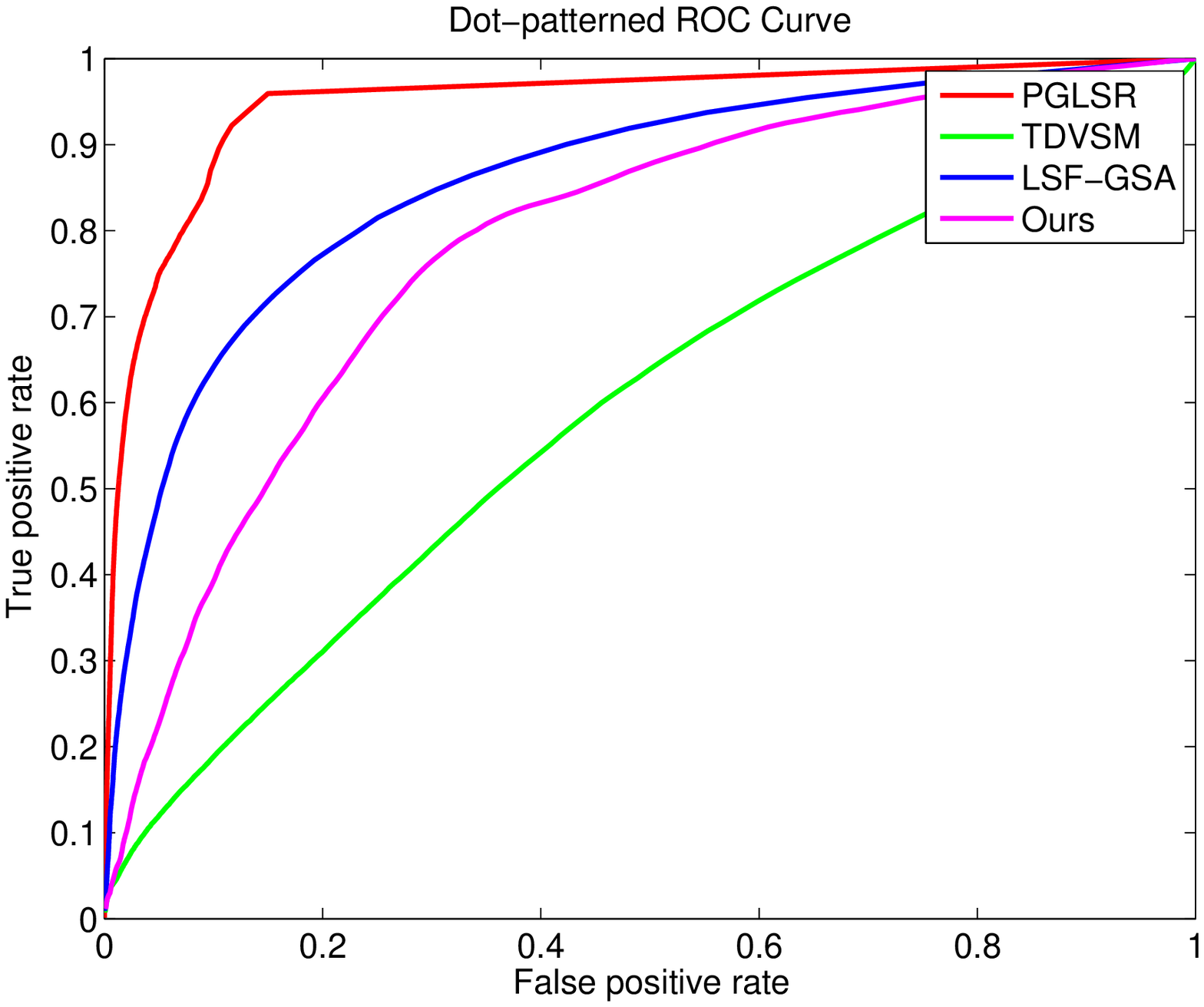}}\quad
  \caption{ROC curves of patterned fabric images compared with other detection methods. (a) ROC curves of the Star-patterned fabric images. (b) ROC curves of the Box-patterned fabric images. (c) ROC curves of the Dot-patterned fabric images.}
\end{figure*}

In addition, some other statistical parameters, such as true positive (TP), true negative (TN), false positive (FP), false negative (FN), are adopted in this paper, originally proposed by Ng.et.al [42]. Based on these parameters, some measure metrics, such as precision, are calculated as:

\begin{equation}
precision = \frac{{TP}}{{TP + FP}}
\end{equation}

and recall is calculated as:

\begin{equation}
recall = \frac{{TP}}{{TP + FN}}
\end{equation}

As shown in Figure 8, we can see that the proposed approach not only provided the highest rate in recall but also provided a balanced performance with respect to precision. Figure 9 takes into account both recall and precision, and adopts the criteria of the F-measure. Additionally, it also demonstrates the effectiveness of our proposed method.

\begin{equation}
F = 2\frac{{precision \cdot recall}}{{precision + recall}}
\end{equation}

\begin{figure*}[htp]
\centering
\subfigure[Star-patterned PR curves]{\includegraphics[width= 0.32\textwidth]{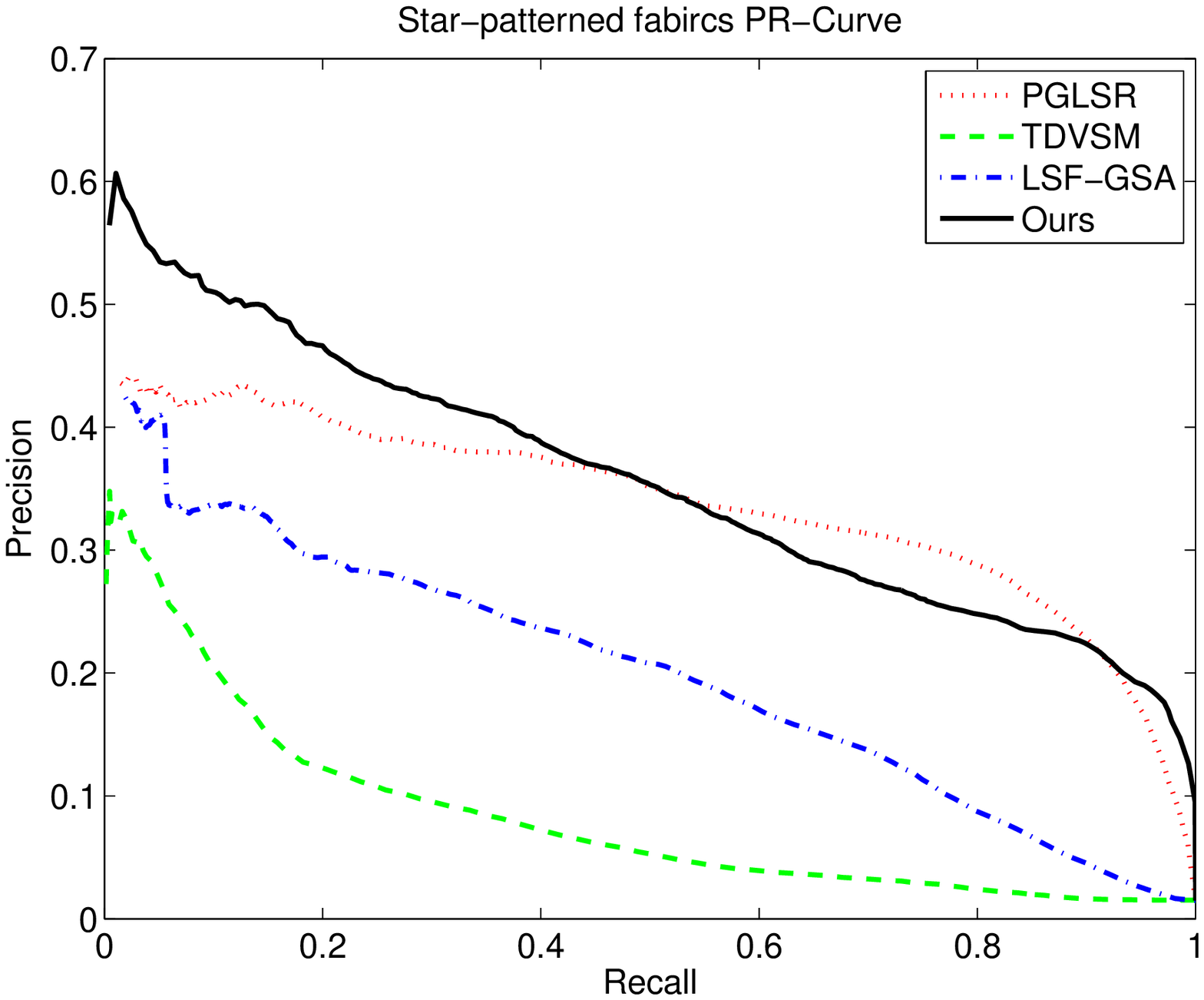}}\quad
\subfigure[Box-patterned PR curves]{\includegraphics[width= 0.32\textwidth]{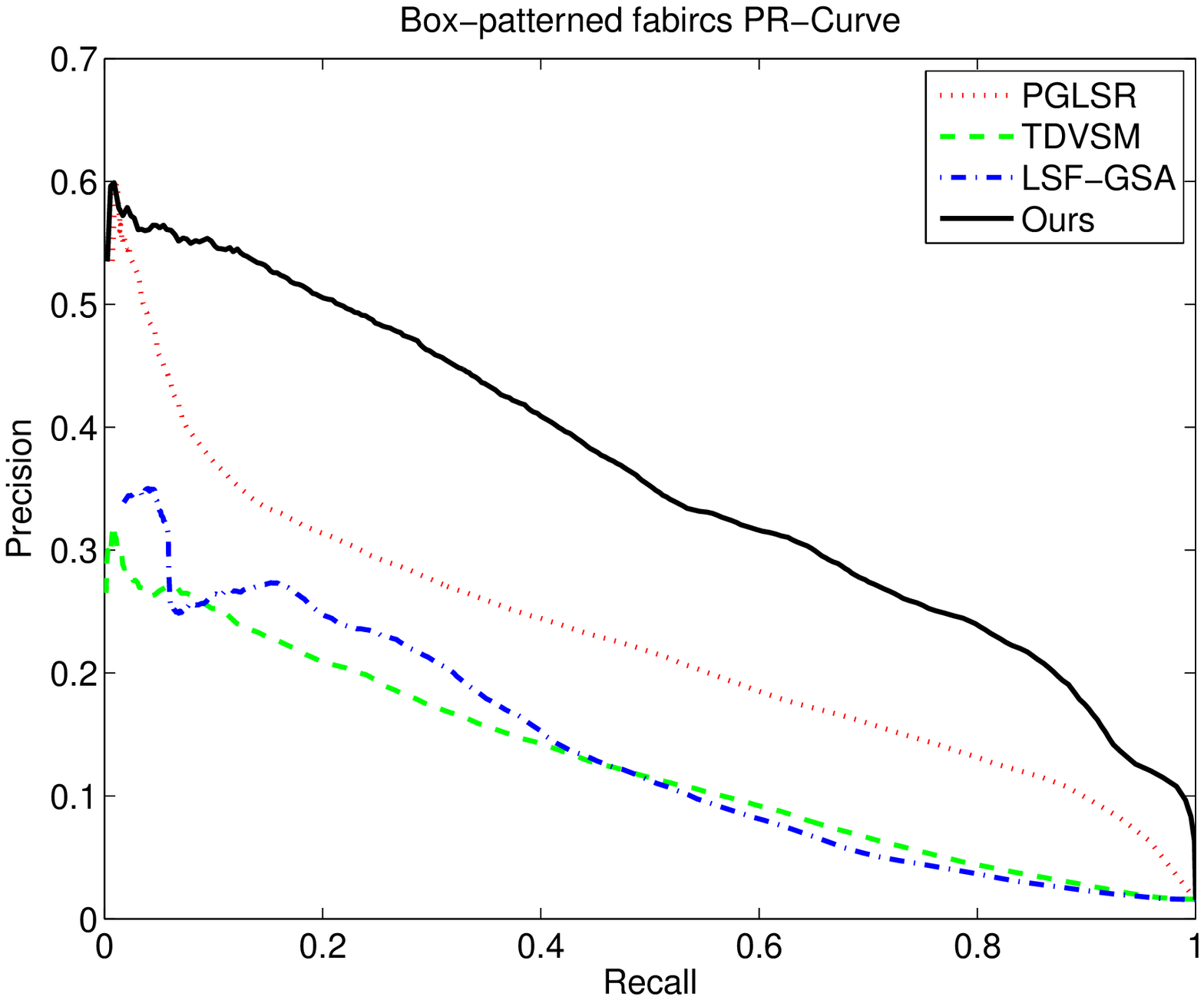}}\quad
\subfigure[Dot-patterned PR curves]{\includegraphics[width= 0.32\textwidth]{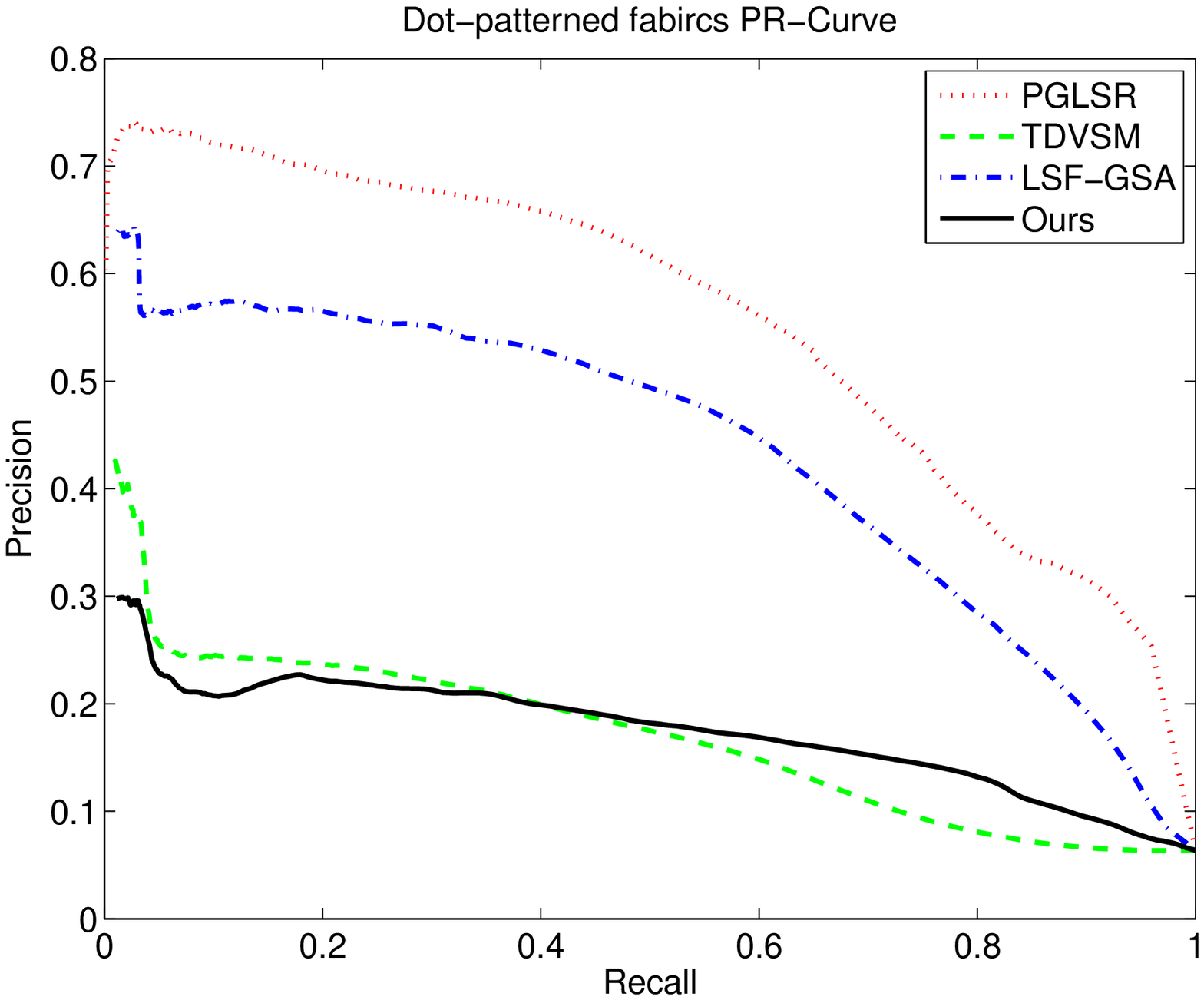}}\quad
  \caption{ROC curves of patterned fabric images compared with other detection methods. (a) ROC curves of star-patterned fabric images. (b) ROC curves of box-patterned fabric images. (c) ROC curves of dot-patterned fabric images.}
\end{figure*}

\begin{figure}[htp]
  \centering
  \includegraphics[width=0.48\textwidth]{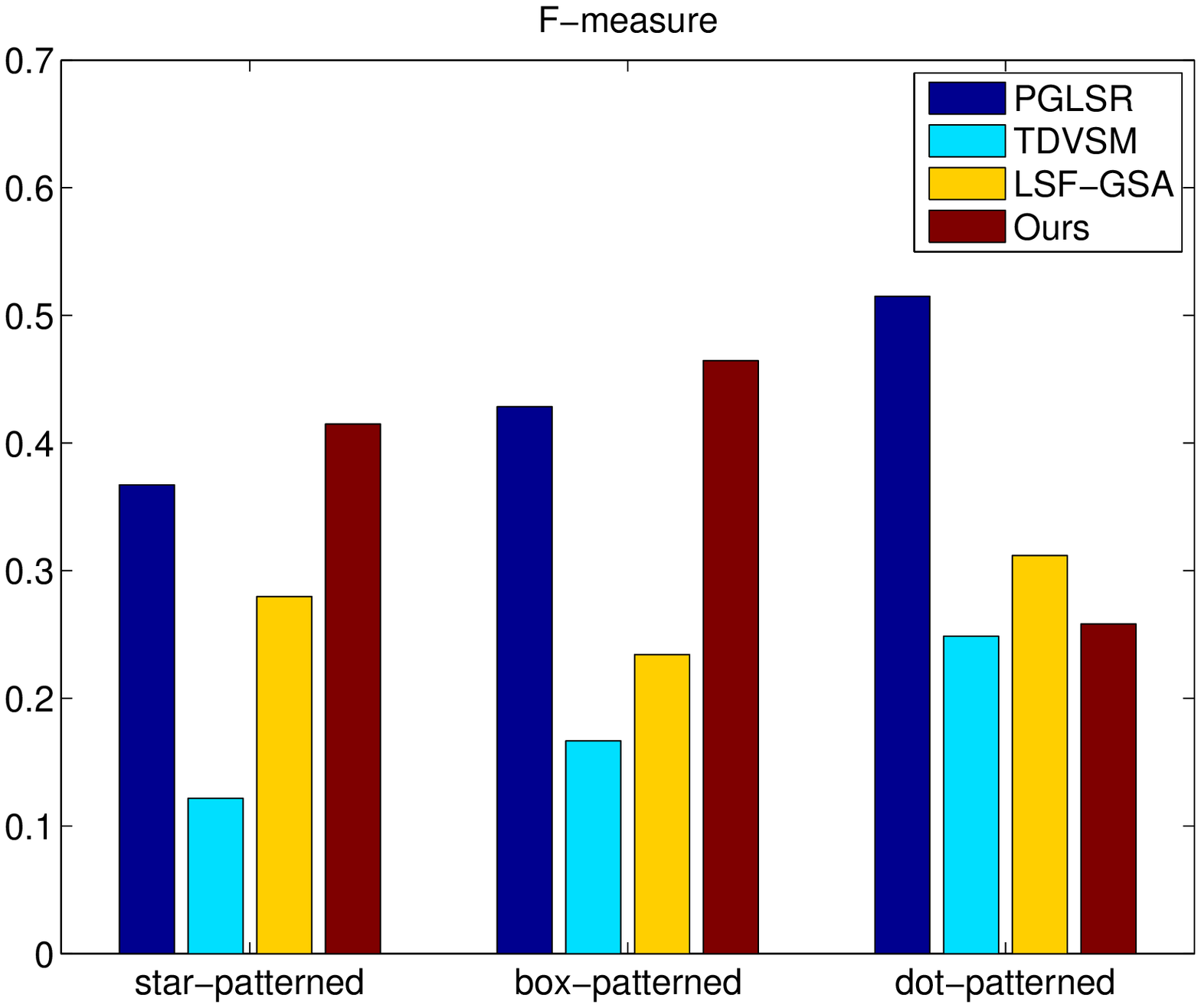}\\
  \caption{The F-measure of our method compared with other methods.}
\end{figure}

\section{Conclusion}
In this paper, we proposed a novel patterned fabric defect algorithm based on GHOG and low-rank decomposition. GHOG descriptor is adopted to extract the features of the patterned fabric image, and the proposed low-rank model is adopted to decompose the fabric image into normal background and defect, respectively. Experimental results demonstrate that the proposed method can efficiently and correctly locate the defect region, and show that our approach outperforms other state-of-the-art methods. In the future, additional theoretical development will be beneficial for defect detection in the glass surface, rail surface, etc.


%

%

\section*{Acknowledgment}
This work was supported by the National Natural Science Foundation of China (No.61202499, No.61379113), Science and technology innovation talent project of Education Department of Henan Province(17HASTIT019), Science and technology leader project of Zhengzhou City(131PLJRC643). In addition, the authors would like to thank Dr. Henry Y.T. Ngan, Industrial Automation Research Laboratory, Dept. of Electrical and Electronic Engineering, The University of Hong Kong, for providing the database of patterned fabric images.

\ifCLASSOPTIONcaptionsoff
  \newpage
\fi

\end{document}